# New Estimation Procedures for PLS Path Modelling

X. Bry

Laboratoire LISE-CEREMADE, Université de Paris IX Dauphine

Email : bryxavier@yahoo.fr

*Abstract :*

*Given R groups of numerical variables $X_1, ... X_R$, we assume that each group is the result of one underlying latent variable, and that all latent variables are bound together through a linear equation system. Moreover, we assume that some explanatory latent variables may interact pairwise in one or more equations. We basically consider PLS Path Modelling's algorithm to estimate both latent variables and the model's coefficients. New « external » estimation schemes are proposed that draw latent variables towards strong group structures in a more flexible way. New « internal » estimation schemes are proposed to enable PLSPM to make good use of variable group complementarity and to deal with interactions. Application examples are given.*

## **Notations**

$I_n$ stands for identity matrix of size *n*. When the matrix's size is unambiguous, we'll simply write it *I*.

Greek lowercase letters ($\alpha, \beta, ...., \lambda, \mu, ...$) stand for scalars.

*X* is a data matrix describing *n* individuals (lines) using variables (columns). This symbol indifferently stands for the variable group coded in the matrix. If variable group *X* contains *J* variables, we then write : $X = \left(x^j\right)_{j=1 to J}$.

*M* is a symmetric regular positive matrix of size *J* used to weigh *X*.

Lowercase *x, y* refer to column-vectors of size *n* as well as to the corresponding variables.

$X_1, ..., X_r, ..., X_R$ are *R* observed variable groups. Group $X_r$ has $J_r$ columns.

$M_r$ is a symmetric regular positive matrix of size $J_r$ used to weigh variable group $X_r$.

*Diag(A,B,C...)* stands for the block-diagonal matrix with diagonal blocks *A, B, C...*

*<X>* stands for the vectorial subspace spanned by variable group *X*.

*F* and $\Phi$ stand for factors built up through linear combination of variables from group *X*.

*v, w* stand for latent variables (to be estimated through factors).

The perpendicular projector onto subspace *E* will be written $\Pi_E$.

*x* being a vector and $E_1, E_2$ two subspaces, the $E_1$-component of $\Pi_{E_1+E_2} x$ will be written $\Pi_{E_1}^{E_2} x$ (Note : the restriction of $\Pi_{E_1}^{E_2} x$ to subspace $E_1+E_2$ is projector onto $E_1$ parallelly to $E_2$).

If not explicitly mentioned, orthogonality will be taken with respect to the canonical euclidian metric *I*.

Scalar product of two vectors *x* and *y* will be written $\langle x|y \rangle$.

Symbol $\propto$ stands for proportionnality of two vectors, e.g. : $x \propto y$.

Standardized variable *x* will be written *st(x)*.

Juxtaposition of matrices *A, B, C...* will be written *[A, B, C ...]*.

A few acronyms :

     PCA: Principal Components Analysis     PLS: partial Least Squares     PLSPM: PLS Path Modelling

N.B. Variables are systematically assumed to have zero mean.

# 1. Introduction

## Conceptual models

Consider a variable group $Y$ describing an aspect of reality (e.g. Health) on $n$ units (e.g. countries) and $R$ explanatory variable groups $X_1, ...X_R$, each pertaining to a theme, i.e. having a clear conceptual unity and a proper role in the explanation of $Y$ (e.g. Wealth, Education...). We may graph the dependancy of $Y$ on $X_1, ...X_R$ as shown on figure 1.

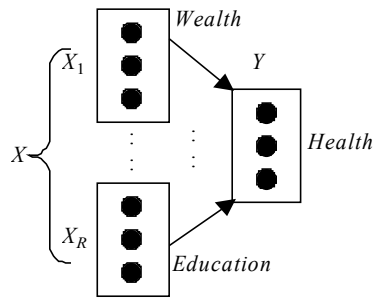

*Figure 1 : Conceptual model*

It should be clear that such a conceptual model shows *no interest* in the global relation between Education and Health over the units, for instance, but aims at knowing whether, *Wealth level etc. remaining unchanged*, a change in Education implies a change in Health. So, arrows in this graph indicate *marginal* (or *partial*) effects, and not global relations between each $X_r$ and $Y$. This analytical approach (the *mutatis mutandis et ceteris paribus* question) is the whole point of such models, so we think estimation strategies should reflect full concern of it.

## Modelling with Latent Variables

• Consider $R$ variable groups $X_1, ..., X_R$. One may assume that underlying each group, there is one latent (unobserved) variable $v_r$, and that these latent variables are linked through a linear model having one or more equations (cf. fig. 2). We shall refer to this model as the *latent model*.

*Figure 2 : Multi-equation latent variables model*

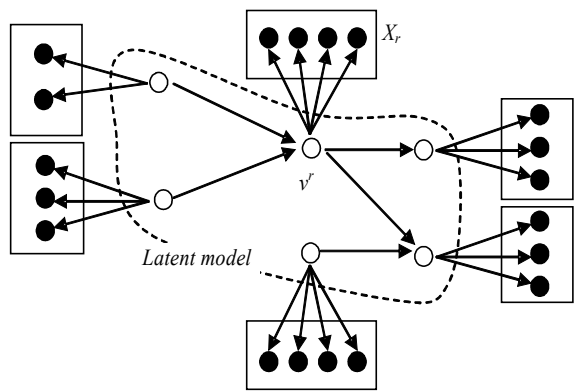

• Consider a single equation latent model, in which $q$ latent variables act upon one observed or latent variable. We shall call the model a *q-predictor-group* one. A 1-predictor group model is essentially different from a multi-predictor group one, in that the effect of a predictor group on the dependant group indicates a *global* relation in the former model, and a *partial* relation (controlling for all other predictor groups) in the latter model.

• Each latent variable underlying a group of observed variables should, as far as possible, represent a group's strong structure, i.e. a component common to as many variables as possible in the group. At the same time, the latent variable set should fulfill, as best as possible, the latent model. It is easily understandable that these two constraints will generally draw the estimation in divergent directions. Therefore, a compromise must be found. It also stands to reason that the estimation of a latent variable with exclusive regard to the observed variables in its group, leaning on global redundancy between them, should make use of simple correlations (as PCA does), whereas its estimation with regard to the latent model, having to deal with effect separation, should use partial correlations. One therefore ends up with two different estimation schemes, and has to make them work hand in hand.

• In multi-predictor group models, the question of collinearity *between* groups arises. We think that, from the moment this type of model has been chosen by the analyst, he should not tolerate collinearity between groups, at least as far as strong within-group structures are concerned : the purpose of the model being to separate the explanatory group effects on the dependant one, these effects must indeed be separable. So, within-group collinearity should be no problem, whereas between-group collinearity should be one.

## 2. PLS Path Modelling's algorithm

• This approach was initially proposed by H. Wold [Wold 85], and improved by Lohmöller [Lohmöller 89]. Wold's algorithm supposed that the sign of the correlation between each dependant latent variable $F_r$ and each of its latent determinants $F_m$ was known *a priori*. This is a costly assumption. Lohmöller's method notably relaxes this hypothesis [Lohmöller 1989][Tenenhaus 1998][Vivien 2002], and carries out estimation without any more information than the conceptual model we have presented here (i.e. mere dependancy arcs between variables). Therefore, let us here present Lohmöller's algorithm.

• Consider $R$ variable groups $X_1, ..., X_r, ... X_R$. All variables are standardized. One makes the following hypotheses:

**H1** : Each group $X_r$ is essentially unidimensional, i.e. is generated by one single latent variable $v_r$. Each variable in the group can thus be written $x_r^k = a_r^k v_r + \varepsilon_r^k$, where $\varepsilon_r^k$ is a centered noise uncorrelated with $v_r$.

**H2** : Latent variables are linked together by such structural relations as : $v_r = \sum_{t \neq r} b_{tr} v_t + \omega_r$, where $\omega_r$ is uncorrelated with the $v_t$'s standing in the righthand side. Some of the $b_{tr}$'s are *a priori* known to be 0. Hypothesis H2 corresponds to the causal relation graph between latent variables. A non-null $b_{tr}$ coefficient means that there is a causal arc oriented from $v_t$ towards $v_r$.

## 2.1. The algorithm

<u>Step 0 (initialization)</u> :

A starting value $F_r(0)$ is determined for each latent variable $v_r$, for instance by equating it to one of the variables of its group. We suggest that one start with the group's first principal component, since it embodies the group's « communality ».

<u>Step $p$</u> :

*Phase 1 (internal estimation of each variable):*

One sets :

$$\Phi_r(p) = st\left(\sum_{t \neq r} c_{tr} F_t(p-1)\right) \quad (1)$$

... where *st(x)* means *standardized x*, and coefficients $c_{tr}$ are computed as follows:

1 – When latent variable $v_r$ is to be explained by variables $\{v_t\}$, coefficients $\{c_{tr}\}$ will be equated to regression coefficients of $F_r$ on $\{F_t\}$ ;

2 - When $v_r$ is an explanatory variable of $v_t$, $c_{tr}$ will be equated to single correlation $\rho(F_t, F_r)$.

*Phase 2 (external estimation of each variable):*

The internal estimation $\Phi_r$ is drawn towards strong correlation structures of group $X_r$ by computing:

$$F_r(p) = st(X_r X_r' \Phi_r(p)) \quad (2)$$

Note that, all variables being standardized, this formula also reads :

$$F_r(p) = st\left(\sum_{x^j \in X_r} \rho(x^j, \Phi_r(p)) x^j\right)$$

<u>End</u> : The algorithm stops when estimations of latent variables have reached required stability.

## 2.2. Discussion

• Algorithmic structure:

Phase 2 (external estimation) of the current step draws each latent variable estimation towards a strong structure of its group using binary correlations between the internally estimated value of the latent variable and all observed variables of its group. We study properties of operator $X_r X_r'$ and generalize it in next section.

Phase 1 (internal estimation) is supposed to bring the estimation of the latent variable closer to the relation it should fulfil with the others. And so it does, to a certain extent. But, according to us, it does not fully comply with the partial correlation logic, and therefore does not make full use of group-complementarity to optimize prediction. This point is developped further below. It seems to us that external and internal estimations, which we try to make meet, each have their own logic :

- External estimation, in order to draw each latent variable towards strong correlation structures within the group, naturally uses single bivariate correlation between the variables in the group.

- Internal estimation, on the contrary, tries to draw the latent variables towards a linear explanatory scheme, i.e. to separate the effects of explanatory groups onto the dependant group. Its logic should therefore use partial correlation.

Convergence of internal estimation may be reached, as well as that of external estimation, but the two limits will generally remain distinct, since external estimation $F_r$ pertains to $< X_r >$, when internal estimation $\Phi_r$ does not *a priori*.

• Internal Estimation scheme:

Let us get back to coefficients $c_{tr}$. Suppose the model has but one equation, predicting latent variable $v_r$ linearly from latent variables $\{ v_t \}$.

The internal estimation of $F_r$ will be:

$$\Phi_r(p) = st\left( \sum_{t \neq r} c_{tr} F_t(p-1) \right)$$

... where coefficients $\{c_{tr}\}$ are regression coefficients of $F_r$ on $\{F_t\}$. These coefficients are partial effects, thus comply with the effect-separation logic.

Now, consider the internal estimation of each $v_t$:

$$\Phi_t(p) = st(c_{rt} F_r(p-1)) \quad \text{where:} \quad c_{rt} = \rho(F_t(p-1), F_r(p-1))$$

When there is but one equation in the model, we see that internal estimation of each predictor is taken as the predicted variable itself. Should there be several equations and $v_t$ appear as predictor of several dependant variables $v_r$, the internal estimation would compute a sum of these dependant variables (externally estimated) wheighed by their global correlation with $F_t$. In any case, the estimation $\Phi_t$ completely ignores the existence of other predictors of $v_r$. The fact that coefficient $c_{rt}$ does not convey any idea of partial relation here seems rather problematic to us.

Let us now develop an alternative approach to external and internal estimation.

## 3. External estimation - Resultants:

### 3.1. Linear resultants

• $X$ being a group of $J$ standardized variables, and $y$ a standardized variable, $XX'y$ will be termed *simple resultant* of $y$ on group $X$, and shorthanded $R_X y$. We have already noted that:

$$XX'y = \sum_{j=1}^{J} \rho(y, x^j) x^j$$

• More generally, let $y$ be a numerical variable and $X$ a group of $J$ numerical variables. Let $M$ be a regular symmetric positive $J \times J$ matrix weighing $X$. We call variable $XMX'y$ : *resultant of y on group X weighed by M*, and shorthand it: $R_{X,M} y$. Matrix $XMX'$ is of course none other than that of $M$-scalar product of observation (row) vectors of $X$. We showed that the resultant can be used to measure the concordance of $y$ with $X$ in two ways : its direction gives the dimension of this concordance in the group's sub-space, and its norm can be used to measure the intensity of the link [Bry 2001].

- Now, let $\alpha \in \mathbf{R}^+$. Matrix $XMX'$ being symmetric positive, it can be powered with $\alpha$. Let us write:

$$R_{X,M}^{\alpha} y = (XMX')^{\alpha} y \qquad (3)$$

and call it the *$\alpha$-degree resultant of y on X,M*.

- In PLSPM, the simple resultant on group $X$ is used to draw a current estimation of a latent variable towards a strong correlation structure of $X$. How does resultant $R_{X,M}^{\alpha} y$ draw $y$ towards a strong structure of $X$, and what structure?

Let $G^k$ be the standardized $k$-th principal component of $X$ weighed by $M$, and $\lambda_k$ be the corresponding eigenvalue. $G^k$ is the eigenvector of $XMX'$ associated with eigenvalue $\lambda_k$. So, it is also the eigenvector of $(XMX')^{\alpha}$ associated with eigenvalue $\lambda_k^{\alpha}$. Therefore, we have:

$$R_{X,M}^{\alpha} y = \sum_k \lambda_k^{\alpha} G^k G^{k'} y = \sum_k \lambda_k^{\alpha} \langle G^k | y \rangle G^k \qquad (4)$$

So, $R_{X,M}^{\alpha} y$ can be viewed as a sum of $y$'s components on $X$'s principal component basis, each component being weighed by the factor's structural force $\lambda_k$ (i.e. the part of total variance it captures) taken with power $\alpha$. As a consequence, $y$ is drawn towards Principal Components of $X$ in proportion of the percentage of variance they capture (put to power $\alpha$), and of their correlation with $y$.

- It is easy to see on (4) that:

  - If $\alpha = 0$: $R_{X,M}^0 y = \sum_k \langle G^k | y \rangle G^k = \Pi_{<X>} y$. No account is taken of correlation structures in group $X$.

  - If $\alpha > 0$, heavier PC's will draw stronger on $y$ (provided it has non-zero correlation with them).

  - If $\alpha \to +\infty$, the heaviest PC with which $y$ has non-zero correlation becomes dominating in the sum, so that $y$ is simply projected onto it.

As a conclusion, continuous parameter $\alpha$ reflects the extent to which we consider correlation structures in $X$. Figure 3 gives a little illustration of how resultants work. Here, we have, for simplicity's sake, a bidimensional $X$ group having PCA eigenvalues $\lambda_1$ and $\lambda_2$ such that $\lambda_1 = 2\lambda_2$ (their magnitudes are figured using a thick line). Variable $y$ is positionned in plane $<X>$, as shown. Then, resultants $R_X^1 y$ to $R_X^3 y$ are computed, $R_X^0 y$ being obviously equal to $y$.

*Figure 3: How resultants work*

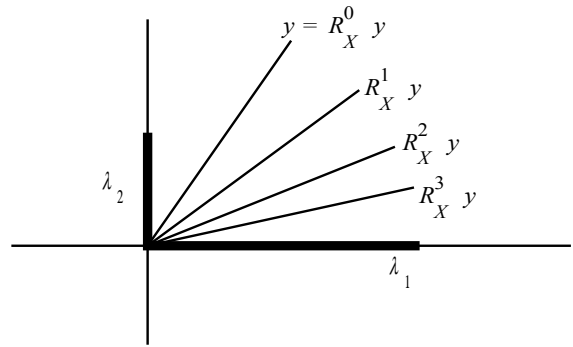

• The purpose of matrix $M$ is also to modulate the account taken of correlation structures in $X$, yet it works a bit differently: less smoothly, in a sense, but allowing to distinguish sub-groups in $X$.

Of course, if we take $M = (X'X)^{-1}$, we end up with:

$$R^\alpha_{X,M} y = \left(X(X'X)^{-1}X'\right)^\alpha y = \left(\Pi_{<X>}\right)^\alpha y = \Pi_{<X>} y \quad \text{(which we will also write: } R_{<X>} y\text{)}$$

... i.e. the same result as when $\alpha = 0$ whatever $M$.

But let us now partition $X$ into $R$ sub-groups $X_1, ..., X_r, ... X_R$ and let $M = Diag(\{(X_r'X_r)^{-1}\}_r)$. Then:

$$R_{X,M} y = \sum_r X_r(X_r'X_r)^{-1}X_r'y = \sum_r \Pi_{<X_r>} y$$

One sees that no account is taken of correlation structures *within* sub-groups, correlations *between* sub-groups still being considered. The immediate application of that is to deal with categorical variables. Suppose we have a group $X$ of $R$ categorical variables. Each such variable $X_r$ will be coded as the dummy variable set of its values. Within this set, correlation structures are irrelevant. So, this set will be regarded as a numerical variable sub-group and one will use $M = Diag(\{(X_r'X_r)^{-1}\}_r)$.

When a numerical variable group $X$ is partitionned into $R$ sub-groups and one wants to take correlation structures into account within each as well as between them, but balance the contribution of groups to the resultant, one may use $M = Diag(\{w_r I_{J_r}\}_r)$, where $J_r$ is the number of variables in group $X_r$ and $w_r$ a suitable weight, for instance the inverse of $X_r$'s first PCA eigenvalue, as in Multiple Factor Analysis [Escofier, Pagès 1990].

### 3.2. Non-linear Resultants

#### 3.2.1. Why may one want to go beyond linear resultants?

Consider a group $X$ of standardized numerical variables, weighed by matrix $I$. We have seen that, for $\alpha > 0$, $R^\alpha_{X,I}$ operator will always draw variable $y$ harder towards stronger PC's (with which it has non-zero correlation), even - and this is the point - if $y$ is very far from them, i.e. these correlations are very low. This can be seen on figure 3: $y$ is closer to the second PC, and yet is drawn towards the first one. Under the hypothesis that group $X$ is fundamentally unidimensional (H1), this is nothing to complain about. But many situations are thoroughly multidimensional. Let us just recall, for History's sake, the great Spearman-Thurstone controversy as to the dimensions of intelligence. Spearman, and those who followed him, were mislaid for 30 years by the prejudice that there was but one factor underlying intellectual aptitudes. Thurstone pointed out the existence, among the psychometric test data, of several positively but weakly correlated variable bundles, and made clear that it was essential that PCs help identify them. Computing several PCs instead of one is a first considerable improvement. The second improvement owed to Thurstone is the rotation he proposed of original PCs so as to make them adjust variable bundles (yet, it may still not be sufficient, as mutually uncorrelated components cannot adjust correcly to correlated bundles).

In our external estimation context, we would like to « draw the estimation towards a close structural direction » (e.g. bundle, if any), still paying some respect to the strength of this structure. It is clear that the linear resultant fails to achieve that. Consider figure 4, showing a group consisting in two positively but weakly correlated variable bundles of equal structural importance.

*Figure 4: Bundles, PCs and resultants*

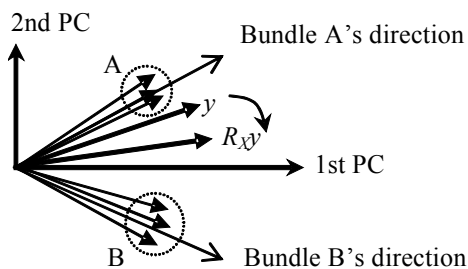

Variable *y* is close to bundle A, and has very little to do with bundle B (which is only weakly correlated with A). Yet, the resultant draws it towards the first PC, and so, towards B. Let us rotate the PCs so as to make each one as close as possible to a bundle, and substitute them for group *X* in the resultant computation. As these rotated PC's capture the same amount of variance (for obvious symmetry reasons) and are uncorrelated, the resultant operator will boil down to the identity matrix, and will leave any variable *y* unchanged (fig. 5). So, *y* is still not drawn towards the closest structural direction.

*Figure 5: Bundles, rotated PC's and resultants*

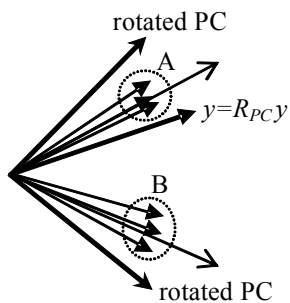

To achieve that, we have to introduce some bonus to « closeness » in the resultant's computation, and this makes it non-linear.

### *3.2.2. Non-linear resultants*

**a) Formulas**

- The simple resultant of *y* on a numerical standardized variable group *X* was calculated as follows:

$$R_X y = \sum_j \langle y | x^j \rangle x^j$$

Taking $\beta \in \mathbf{R}^+$, we may introduce a bonus to closeness by calculating instead:

$$\sum_j \left| \langle y | x^j \rangle \right|^\beta \langle y | x^j \rangle x^j$$

It may be sufficient, in practice, to take $\beta = 2k$, where *k* is a natural integer, which we shall refer to as the resultant's *order*. Then, we write:

$$S_{X,k}(y) = \sum_j \langle y | x^j \rangle^{2k+1} x^j$$

• Let us generalize the previous situation by considering a numerical variable group $X$ partitionned into $R$ sub-groups $X_1, ..., X_r, ... X_R$ and let $M = Diag(\{(X_r'X_r)^{-1}\}_r)$. The linear resultant was:

$$R_{X,M}\, y = \sum_r \Pi_{<X_r>}\, y$$

Now, if we introduce our bonus to closeness in the same way as above, we have:

$$S_{X,k}(y) = \sum_r \cos^{2k}(y, <X_r>)\, \Pi_{<X_r>}\, y = \sum_r \left\| \Pi_{<X_r>}\, y \right\|^{2k} \Pi_{<X_r>}\, y \quad (5)$$

This formula generalizes the previous one and allows us to deal with categorical variables.

It can obviously be written:

$$S_{X,k}(y) = X M_{X,y,k} X'\, y \quad (6)$$

... where $M_{X,y,k} = Diag\left( \left\{ \left\| \Pi_{<X_r>}\, y \right\|^{2k} (X_r'X_r)^{-1} \right\}_r \right)$ is a symmetric positive matrix including the « bonus to closeness » effect and thus, depending on $y$ (hence the non-linearity). Matrix $M_{X,y,k}$ is a local euclidian metric matrix. So, matrix $S_{X,y,k} = X M_{X,y,k} X'$ is a local resultant operator. Just as linear resultant operators, it can be put to any positive power $\alpha \in \mathbf{R}^+$:

$$S^{\alpha}_{X,y,k} = \left( X M_{X,y,k} X' \right)^{\alpha} \quad (7)$$

As with linear resultants, $\alpha$ can be interpreted as the degree of account taken of structures in $X$.

**b) Behaviour**

When $\alpha = 0$, we get the orthogonal projection onto $<X>$. Now, let us take $\alpha > 0$:

When $k = 0$, we get the linear resultant back, the bonus to closeness being null.

When $k > 0$, variable subgroups spanning subspaces closer to $y$ are given more weight.

When $k \to \infty$, the variable subgroup spanning the closest subspace to $y$ is dominating: $S^{\alpha}_{X,k}(y)$ is colinear to $y$'s projection onto this subspace. When sub-groups are reduced to single variables, $S^{\alpha}_{X,k}(y)$ ends up being colinear to the variable best correlated with $y$.

Let us illustrate this with an example. Using a random number generator, we computed a variable group $X$ consisting in 2 numerical variable bundles ($A$ and $B$) approximately making a $\pi/4$ angle. Bundle $A$ contains 4 variables ($a^1, ... a^4$) obtained by adding a little random noise to the same variable. Bundle $B$ only contains 2 variables ($b^1, b^2$), generated in the same way. Bundle $B$ is thus « lighter » than $A$. Then, several $y^j$ variables are generated through linear combination of variables in $X$. Finally, we computed non-linear resultants of $y^j$'s on $X$ with $k$ ranging from 0 to 6. Resultant $S^1_{X,k}(y)$ will be shorthanded $S_k y$.

Figure 6 shows what becomes of a variable ($y^7$) located inbetween bundles $A$ and $B$, but closer to $B$, according to the $k$ value (all variables are projected onto $X$'s first PCA plane).

Figure 7 shows nl-resultants for all variables and *k*-values 0 (linear resultant) and 6 (furiously non-linear resultant). It is easy to notice that $S_6$ resultants are grouped in the bundles' neighbourhoods, in contrast with $S_0$ ones. Variables $y^4$, $y^5$ and $y^7$ have been drawn by $S_6$ towards bundle *B*, whereas $y^1$, $y^2$, $y^3$ and $y^6$ have been drawn towards bundle *A*.

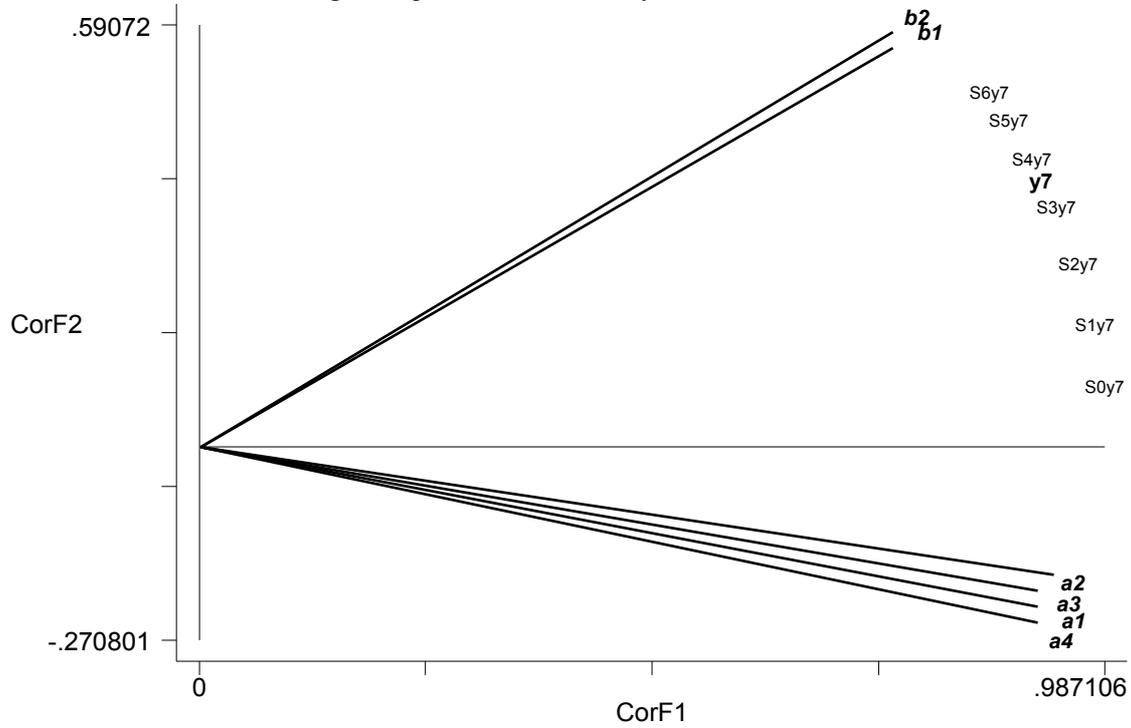

*Figure 6: possible attractions of an « inbetween » variable*

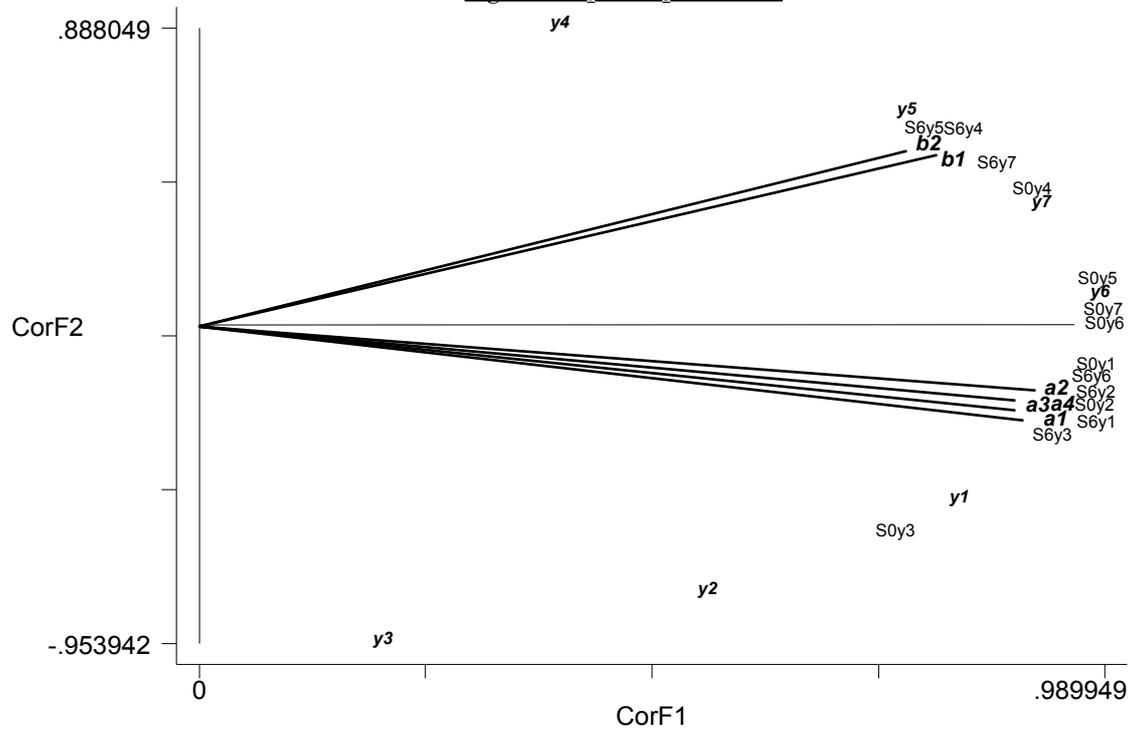

*Figure 7: $S_0$ and $S_6$ resultants*

**c) Link with the quartimax rotation**

A link may be established between 1st order non linear resultant and *quartimax* rotation. This rotation aims at drawing a set of $H$ orthogonal standardized factors closer to $H$ variable bundles. This method has been derived by several authors ([Ferguson 1954] [Carroll 1953] [Neuhaus & Wrigley 1954] [Saunders 1960]) from distinct but equivalent criteria. For instance, it can be derived from the following program:

$$\underset{\substack{F^1,...,F^H \\ \text{orthogonal} \\ \text{standardized}}}{Max} \sum_{h=1}^{H} \sum_{j} \cos^4(x^j, F^h)$$

Keeping the heuristic base of the program, we can extend it to any even power greater than 4. Thus, for $k \geq 2$, we get:

$$\underset{\substack{F^1,...,F^H \\ \text{orthogonal} \\ \text{standardized}}}{Max} \sum_{h=1}^{H} \sum_{j} \cos^{2k}(x^j, F^h)$$

This program can be rewritten:

$$\underset{\substack{F^1,...,F^H \\ \text{orthogonal} \\ \text{standardized}}}{Max} \sum_{h=1}^{H} \left\langle \sum_{j} \cos^{2k-1}(x^j, F^h) x^j \middle| F^h \right\rangle$$

That is:

$$\underset{\substack{F^1,...,F^H \\ \text{orthogonal} \\ \text{standardized}}}{Max} \sum_{h=1}^{H} \left\langle S_{X,k-1}(F^h) \middle| F^h \right\rangle$$

In every scalar product $<S_{X,k-1}(F^h) | F^h>$, two elements are taken into account: the correlation of the factor with its non-linear resultant, and the norm of that resultant. The non-linear resultant drawing a factor $F$ towards a « strong and close » structure, the factor itself will be all the closer to this structure as it is correlated to the resultant. On the other hand, the resultant's norm will be all the greater as $F$ is close to the structure. Thus, the criterium maximized by the program is straightforward to interpret.

# 4. Internal estimation

We are now going to set up an alternative internal estimation scheme. Let us take back notations from §2. External estimates of latent variables will be referred to as *factors*.

## 4.1. Latent model without interaction effects

### 4.1.1. Single equation model

Take latent model: $v_r = \sum_{t \neq r} b_{tr} v_t + \varepsilon_r$. External estimates of $v_r$, $v_t$ computed at step $p$-1 are $F_r(p\text{-}1)$, $F_t(p\text{-}1)$. Indices $r$ and $t$ will refer to dependant and explanatory variables respectively. Let $F_{-t}$ stand for the the set of all explanatory factors except $F_t$.

**Estimation formulas:**

• To estimate $v_t$, we have to take into account all other predictors of $v_r$, and shall take advantage of the whole prediction potential of group $X_t$. So, we regress $F_r(p\text{-}1)$ onto $\{X_t, F_{-t}(p\text{-}1)\}$, and take the component on $X_t$ of the prediction. So, we have:

$$\Phi_t(p) = st\left(\Pi_{\langle X_t \rangle}^{\langle F_{-t}(p-1) \rangle} F_r(p-1)\right) \quad (8)$$

• We could keep estimating $v_r$ internally using Lohmöller's method, i.e. formula (1), $\{c_{tr}\}$ being the coefficients of $F_r(p\text{-}1)$'s regression upon $\{F_t(p\text{-}1)\}$. Geometrically speaking, we would then have:

$$\Phi_r(p) = st\left(\Pi_{\langle \{F_t(p-1)\}_t \rangle} F_r(p-1)\right)$$

In fact, we would like, just as for the explanatory variables, to get an internal estimation of $v_r$ that is in $\langle X_r \rangle$, in order to be able to skip external estimation.

So, we shall take:

$$\Phi_r(p) = st\left(\Pi_{\langle X_r \rangle} \Pi_{\langle \{F_t(p-1)\}_t \rangle} F_r(p-1)\right) \quad (9)$$

**Properties:**

• Formula (8) uses partial relations between dependant and explanatory variables.

• As each variable $v_r$'s internal estimation now pertains to subspace $\langle X_r \rangle$, it becomes possible to skip external estimation in the PLSPM algorithm. The roles of internal and external estimations are well separated: internal estimation's purpose is to maximize latent model adjustment, whereas external estimation's purpose is to draw estimations towards groups' strong structures.

• Notice that (8) maximises coefficient $R^2$ over $\langle X_t \rangle$, given $F_{-t}$. If we iterate (8) to internally estimate in turn all explanatory variables $v_t$, $R^2$ increases throughout the process. Formula (9) can only increase $R^2$ too. So, if we skip external estimation, $R^2$ can only increase throughout the PLSPM algorithm.

• Now, let us regress $F_r(p\text{-}1)$ onto all explanatory groups, i.e. $X$, where $X = [\{X_t\}_t]$, and let $F_t^r(p-1)$ be the $\langle X_t \rangle$-component of the prediction. If we replace every current explanatory factor $F_t(p\text{-}1)$ by $F_t^r(p-1)$, then regressing onto $\{X_t, F_{-t}(p\text{-}1)\}$ amounts to the same as regressing on $X$, and therefore, (9) yields:

$$\Phi_t(p) = st\left(F_t^r(p-1)\right)$$

and of course, (8) gives: $\Phi_r(p) = F_r(p-1)$. So, we have a fixed point of the algorithm.

• If there is but one explanatory group $X_t$, and we skip external estimation, the algorithm will perform canonical correlation analysis. Indeed, once stability is reached, the estimated variables verify the following characteristic equations:

$$F_t(\infty) = st\left(\Pi_{<X_t>} F_r(\infty)\right)$$

and $\quad F_r(\infty) = st\left(\Pi_{<X_r>}\Pi_{<F_t(\infty)>} F_r(\infty)\right) \propto st\left(\Pi_{<X_r>} F_t(\infty)\right)$

If we add up standard external estimation using simple resultants, we get rank 1 factors of PLS regression.

But again, the thing to focus on is the use of partial relations to predictors in the estimation process when there are several predictor groups.

**Illustration of the main difference with Lohmöller's procedure**

Consider the case reported on figure 8. Here, we have a dependant group reduced to a single variable $y$, to be predicted from two groups $X_1$ and $X_2$. Explanatory group $X_1$ contains standardized variables $a$ and $b$, which are supposed to be uncorrelated, while $X_2$ merely contains $c$. The only latent variable to be estimated is $v_1$ in group $X_1$. The dependant variable $y$ pertains to plane $<a,c>$, and is such that its orthogonal projection onto $<X_1>$ is colinear to $b$ (i.e. $y = <a,c> \cap <b,<X_1>^\perp>$). The dimension in $X_1$ that is the most useful, together with $c$, to predict $y$ is therefore $a$.

Whatever the initial $F_1$ value, Lohmöller's procedure will internally estimate $v_1$ as $\Phi_1 = y$. Then, external estimation will replace it with $F_1 = X_1 X_1' y = \Pi_{<X_1>} y = b$ (once standardized). We can see that we have reached stability.

*Figure 8*

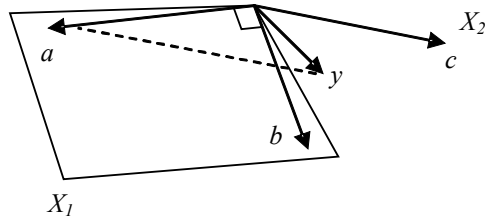

In contrast, the procedure we suggest, projecting $y$ onto $<X_1>$ parallelly to $c$, will find $\Phi_1 = a$. Then, external estimation will replace it with $F_1 = X_1 X_1' a = \Pi_{<X_1>} a = a$. Here also, we have reached stability.

**Collinearity problems:**

• Let $a$ be a linear combination of $F_{-t}$ factors. We have: $\Pi_{<X_t>}^{<F_{-t}>} a = 0$. So, (8) ensures that $\Phi_t$ is as far as possible from collinearity with $F_{-t}$. Of course, $\Pi_{<X_t>}^{<F_{-t}>} F_r$ is uniquely determined only if $<X_t> \cap <F_{-t}> = 0$. This will be the case in practice, provided group $X_t$ is not too large (with respect to the number of observations). If it is, one possibility consists in reducing its dimension through prior PCA. We regard it as a good one, because really weak dimensions should anyway be discarded, and using a subspace of $<X_t>$, we keep $\Phi_t$ in $<X_t>$.

A remaining collinearity problem between $<X_t>$ and $<F_{-t}>$, once the weaker dimensions in $<X_t>$ removed, would indicate that $X_t$ shares strong explanatory dimensions with the other explanatory groups. This is a problem, but less with the method than with the conceptual model: this means indeed that the partial effects of explanatory groups upon the dependant one can theoretically *not* be separated, which violates the fundamental assumption of our analytical approach. Under such circumstances, how could one think of an explanatory model?

● When there is some collinearity within $X_t$, $\Pi^{<F_{-t}>}_{<X_t>} F_r$ cannot be expressed uniquely in terms of $x_t^j$ variables. Nevertheless, if $<X_t> \cap <F_{-t}> = 0$, $\Pi^{<F_{-t}>}_{<X_t>} F_r$ exists and is unique, which is enough for our estimation method (it could be computed using $X_t$'s PCs). So collinearity within $X_t$ causes no real problem.

*4.1.2. Multi-equation model*

● Here, we suppose that the model contains several equations. A latent variable $v_r$ can be the dependant one in some equations, and explanatory in others. Supposing $v_r$ intervenes in $Q$ equations. We may apply formulas (8) and (9) to estimate separately $v_r$ in each equation. Equation $q$ leads to internal estimation $\Phi_r^q(p)$.

How can we synthesize a unique internal estimation from all these separate estimations? Let $\Omega_r(p) = \{\Phi_r^q(p)\}_q$ and $\alpha > 0$. A simple and natural way is to set $\Phi_r(p) = R^\alpha_{\Omega_r(p),I} F_r(p-1)$, for instance.

● In what follows, we will refer to this PLSPM algorithm as the *Thematic Components PLS Path Modelling* (shorthanded TCPM). The reason for this being that the idea of projecting the dependant variable onto each explanatory group parallelly to all other explanatory factors was first developed in an algorithm called *Thematic Components Analysis*, dealing with a single equation model [Bry 2003].

*4.1.3. Application example: the Senegalese presidential election of 2000*

The election of the Senegalese president has two ballots. The two candidates who get the highest scores in the first ballot are the only ones to compete in the second ballot. The winner is the one who wins the relative majority in the second ballot.

**The data : (cf. appendix A)**

Senegal is divided into 30 departments. We shall try to relate the departemental scores of the candidates to economic, social and cultural characteristics of the departments, using the conceptual model shown on figure 9. The *rationale* behind this model, is that :

1) Concerning the first ballot, the economic and social situation should be an important factor of political choice. Besides, in a given socio-economic situation, cultural considerations such a ethnic and religious background may still cause differences in the vote.

2) The results of the second ballot are mainly determined by those of the first, through a strong vote-transfer mechanism.

*Figure 9 : conceptual model for the senegalese elections*

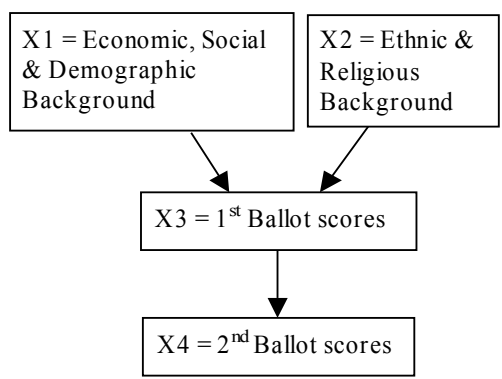

The model has two parts :      (a) X3 = f(X1 , X2)      and      (b) X4 = f(X3).

The latent variable in group *Xk* will be written *Fk*.

The variables used in this analysis are :

*Group X1 (economic social and demographic background)*:

**NHPI :** Normalized Human Poverty Index (a compound measure of educational, sanitary and life conditions indicators)

**PctAgriInc** : Proportion of the global income that is made in Agriculture

**IncActivePers** : Average Income of an active person

**ActivePop** : proportion of the active persons in the population.

**Scol** : Gross Enrolment Ratio

**Malnutrition** : Malnutrition rate.

**DrinkWater** : Proportion of population having access to drinkable water.

**Rural**, **Urban** : Percentages of rural and urban populations.

**PopDensity** : Population Density

**HouseholdSize** : Average number of persons in a household

**Pop0_14**, **Pop15_60**, **PopOver60** : Proportions of population aged 0 to 14, 15 to 60, and over 60.

**WIndep**, **WPublic**, **Wprivate**, **WApprentice** : Proportions of employed population working as independant worker, Public Sector Salaried worker, Private Sector Salaried worker, and Apprentice.

**OEmployed**, **OUnemployed**, **OStudent**, **OHouseWife**, **ORetired** : Proportions of population being : employed, unemployed, student, housewife, retired.

**ASPrim**, **ASSec**, **ASTer** : Activity Sectors ; respectively Primary, Secondary and Tertiary.

*Group X2 (ethnic and religious background) :*

**Wolof**, **sereer**, **joola**, **pulaar**, **manding** : Percentages of the main ethnic groups.

**Moslims** : Percentage of moslim population.

*Group X3 (1st ballot scores) :*

**Thiam1**, **Niasse1**, **Ka1**, **Wade1**, **Dieye1**, **Sock1**, **Fall1**, **Diouf1**, **Abstention1** : Departmental scores of candidates, and abstention. Every candidate score is calculated as number of votes for the candidate over number of electors on the official list.

*Group X4 (2nd ballot scores) :*

**Diouf2**, **Wade2**, **Abstention2** .

We must mention that candidates Thiam, Dieye, Sock and Fall most generally had very low scores at the first ballot (below 1%), and therefore had very little weight in the coalitions that formed between the two ballots. Attention should thus be focussed on the other candidates (Wade, Diouf, Niasse, Ka).

**TCPM results:**

We thought more careful to first investigate equations (a) and (b) separately, and check the closeness of estimations for the common latent variable *F3*, before launching the joint estimation of the whole 2-equation system. Stability (changes lower than 1/1000) was always reached in less than 10 iterations.

*Separate estimations of equations (a) and (b)*

We successively used resultants $S_0$, $S_1$, $S_2$, $S_3$, $S_4$ to see whether adjustment could be improved. Table 1 gives, for each *S*-choice, the R² coefficient for equations (a) and (b) estimated separately, as well as the correlation between the two estimations of the *F3* factor.

*Table 1: Separate estimation adjustment quality according to S-choice*

| *S operator* | *R² equation (a)* | *R² equation (b)* | *Corr (F3(a),F3(b))* |
|---|---|---|---|
| $S_0$ | .686 | .912 | .897 |
| **$S_1$** | **.711** | **.925** | **.913** |
| $S_2$ | .669 | .895 | .914 |
| $S_3$ | .658 | .864 | .849 |
| $S_4$ | .653 | .821 | .767 |

All in all, the choice of $S_1$ seems best, as it gives better adjustment of both equations, and (almost) the best correlation between F3 estimations. Such a good correlation will entitle us to proceed later on to a joint estimation of both equations.

**Interpretation of equation (a) estimated on its own :**

Comparing the factors given by the different *S* order options, we noticed that $S_1,... S_4$ give out very close results (correlations between estimations of the « same » factor ranging from 0.97 to 1), but that there are more important differences between the results given by $S_0$ and $S_1$, especially in group X2, as shown below.

| *Factor* | *F1* | *F2* | *F3* |
|---|---|---|---|
| correlation between $S_0$- and $S_1$-estimations | .994 | .491 | .815 |

Considering this, we found important to interpret results in the $S_0$ and $S_1$ cases.

● *F3* factor was regressed on *F1* and *F2* , which gave the following results :

$S = S_0$:  R² = .686

| Explanatory factor → | F1 | F2 |
|---|---|---|
| Coefficient | .632 | -.434 |
| P-value[1] | .000 | .000 |

---

[1] Critical P. values should not be used for inference, but only be taken as a descriptive indicator.

$S = S_1$:  R² = .711

|  | Explanatory factor → | F1 | F2 |
|---|---|---|---|
|  | Coefficient | .776 | .304 |
|  | P-value[1] | .000 | .007 |

● Interpretation of *F3* :

| F3-Correlations | Thiam1 | Niasse1 | Ka1 | Wade1 | Dieye1 | Sock1 | Fall1 | Diouf1 | Abst.1 |
|---|---|---|---|---|---|---|---|---|---|
| $S = S_0$: | .226 | **.467** | **-.596** | **.730** | -.633 | -.542 | .334 | **-.553** | -.383 |
| $S = S_1$: | .148 | .071 | **-.416** | **.975** | -.277 | -.243 | .150 | **-.729** | -.263 |

Let us consider the candidates having non-skinny scores.

$S = S_0$: Factor F3 opposes MM. Wade and Niasse to MM. Diouf and Ka. Note that after the first ballot, candidates Diouf and Ka formed a conservative coalition, whereas candidates Wade and Niasse formed a coalition to « change » (« Sopi », meaning « change » in wolof, was their slogan).

$S = S_1$: Here, factor *F3* opposes M. Wade to M. Diouf, and to a modest extent, M. Ka. The correlation with M. Wade's score is much higher, but M. Niasse has been lost on the way.

● Interpretation of *F1*:

| F1-Correlations | NHPI | PctAgriInc | IncActivePers | ActivePop | Scol | Malnutrition | DrinkWater | Rural | Urban | PopDensity | HouseholdSize | Pop0_14 | Pop15_60 |
|---|---|---|---|---|---|---|---|---|---|---|---|---|---|
| $S = S_0$: | **-.937** | -.437 | .713 | -.294 | .725 | -.021 | .688 | **-.956** | **.956** | .767 | -.086 | -.720 | .732 |
| $S = S_1$: | **-.942** | -.451 | .730 | -.285 | .672 | .045 | .714 | **-.960** | **.960** | .781 | -.050 | -.733 | .774 |

|  | PopOver60 | WIndep | WPublic | Wprivate | WApprentice | OEmployed | OUnemployed | OStudent | OHouseWife | ORetired | ASPrim | ASSec | ASTer |
|---|---|---|---|---|---|---|---|---|---|---|---|---|---|
| $S = S_0$: | -.320 | -.564 | **.931** | **.963** | -.700 | **-.834** | **.870** | .607 | .513 | -.097 | **-.956** | **.914** | **.940** |
| $S = S_1$: | -.385 | -.615 | **.944** | **.968** | -.656 | **-.815** | **.885** | .543 | .543 | -.141 | **-.955** | **.906** | **.942** |

From these correlations, it is clear that in both cases, F1 opposes urban departments (high values) to rural ones (low values). Urban departments have a higher population density, are better provided with industry and services and are relatively rich, whereas rural ones merely depend on agriculture and are very poor.

● Interpretation of *F2*:

| F2-Correlations | Wolof | Sereer | Joola | Pulaar | Manding | Moslims |
|---|---|---|---|---|---|---|
| $S = S_0$: | -.469 | -.628 | -.271 | **.907** | .428 | .316 |
| $S = S_1$: | -.225 | -.011 | **.949** | -.534 | .076 | **-.878** |

$S = S_0$: Factor F2 is highly correlated with the presence of the Pulaar ethnic group.

$S = S_1$: Here, factor F2 is highly correlated positively with the presence of the Joola ethnic group and negatively with the proportion of moslims in the population. Recall that the great majority of the Joola people lives in the south deparments and is christian, and that most christians are Joolas, too, hence an interdepartmental correlation of -.839 between the percentage of Joolas and that of moslims.

In this case, the choice of *S* has heavy consequences: the computed factors do not quite seem to point to the same phenomenon, and will not lead to identical models.

● We end up with the following models, relating standardized factors:

$S = S_0$:            *F3* =  .632  *F1* -  .434  *F2*       ($R^2$ = .686)

$S = S_1$:            *F3* =  .776  *F1* +  .304  *F2*       ($R^2$ = .711)

If we select the variable best correlated with the factor in each group, we have:

$S = S_0$:          *Wade1* = .204  *PctUrban* -  .053  *Pulaar* +  .119       ($R^2$ = .595)

            (P)      (0.000)           (0.193)            (.000)

$S = S_1$:          *Wade1* = .215  *PctUrban* +  .107  *Joola* +  .094       ($R^2$ = .646)

            (P)      (0.000)           (0.021)            (.000)

Note that regression of *Wade1* onto *PctUrban* alone gives $R^2$ = .568. The *Pulaar* factor does not seem to have a significant role in the model of *Wade1*, whereas the *Joola* factor has, and provides a better prediction.

If we try to model the variable the most negatively correlated with F3 in the case $S = S_0$, i.e. M. *Ka*'s score, we find:

            *Ka1* = .000  *PctUrban*  +  .117  *Pulaar*   +.015 ($R^2$ = .395)

            (P)    (0.994)              (0.000)         (.288)

So, the use of $S_1$ and $S_0$ has directed us towards two different phenomenons: the urban factor and the Joola region's bonus in the Wade vote ($S_1$), and the Pulaar factor in the Ka vote ($S_0$). But the first phenomenon is globally more important, and was more clearly set out.

**Interpretation of equation (b) estimated on its own :**

Here, the choice of *S* does not lead to very different results. We selected $S = S_1$, since it provides the best adjusted latent model.

● Interpretation of *F3* :

| Correlations | Thiam1 | Niasse1 | Ka1 | Wade1 | Dieye1 | Sock1 | Fall1 | Diouf1 | Abst.1 |
|---|---|---|---|---|---|---|---|---|---|
| F3 | .307 | .230 | **-.670** | **.865** | -.275 | -.130 | .283 | **-.803** | -.046 |

F3 opposes the Wade liberal vote to the conservative socialist vote represented by MM. Diouf and Ka.

● Interpretation of *F4* :

| Correlations | Wade2 | Diouf2 | Abst.2 |
|---|---|---|---|
| F4 | **.921** | **-.955** | -.126 |

F4 opposes final Wade and Diouf votes.

- The estimated latent model is:

$$F4 = .962\ F3 \qquad (R^2 = .925)$$

Selecting the observed scores best correlated with the factors and relevant with political orientations, we easily get to the following model of the final Diouf score, expressed as a function of his former score and that of his 2nd ballot ally, M. Ka:

$$Diouf2 = .944\ Diouf1 + .786\ Ka1 - .025 \qquad (R^2 = .935)$$
$$(P)\quad\ (.000) \qquad\quad (.000) \qquad (.174)$$

It is also possible to model the final Wade score in the same way, with nearly equivalent quality.

*Joint estimation of equations (a) and (b):*

We know enough, by now, about the two « conceptual equations » (*economic & cultural* → *vote1* and *vote1* → *vote2*) to try and match them through the joint estimation process.

*Table 2: Joint estimation adjustment quality according to S-choice*

| *S operator* | *R² equation (a)* | *R² equation (b)* |
|---|---|---|
| $S_0$ | .655 | .861 |
| $S_1$ | .648 | .843 |
| $S_2$ | .642 | .746 |
| $S_3$ | .655 | .713 |
| $S_4$ | .662 | .684 |

As shown on table 2, the best latent model global adjustment quality was obtained for $S = S_0$, but results given by $S_0$ and $S_1$ are rather close. Besides, when correlating F2 estimations, one can notice a drastic change in F2 when one leaves $S_0$ or $S_1$ for $S_2$ or a higher order $S$. So, we may feel important to present $S_0$- and $S_2$-estimations.

**Factor interpretation :**

F1 has exactly the same interpretation as in the separate estimation (*rural / urban*). So has F4 (*Wade2 / Diouf2*). As for F2 and F3, $S_0$ leads to the same phenomenon as in the separate estimation (*Pulaar* factor in the *Ka* vote), whereas $S_2$ leads to the phenomenon outlined by $S_1$ in the separate estimation (*urban* factor and *Joola* bonus for *Wade*).

**Final models :**

We end up with two possible models, of unequal adjustment quality, but pointing out different and equally interesting phenomenons. If we select the one or two best correlated variables with each factor, we get:

$S_0 \to$ Model 1:

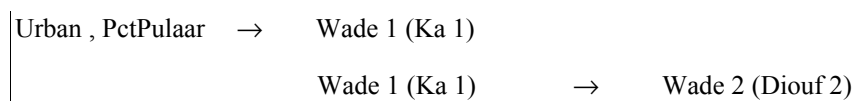

Urban , PctPulaar   →   Wade 1 (Ka 1)

Wade 1 (Ka 1)   →   Wade 2 (Diouf 2)

$S_2 \to$ Model 2:

$$\begin{array}{l} \text{Urban , PctJoola (PctMoslims)} \quad \to \quad \text{Wade 1 (Diouf 1)} \\ \phantom{\text{Urban , PctJoola (PctMoslims)} \quad \to \quad} \text{Wade 1 (Diouf 1)} \quad \to \quad \text{Diouf 2 (Wade 2)} \end{array}$$

Combining these two models gives our final path model of the contest (fig. 10). All coefficients have P-value below .01 except for the Joola effect in Wade1's model (P = .02).

*Figure 10: final path model of the electoral contest*

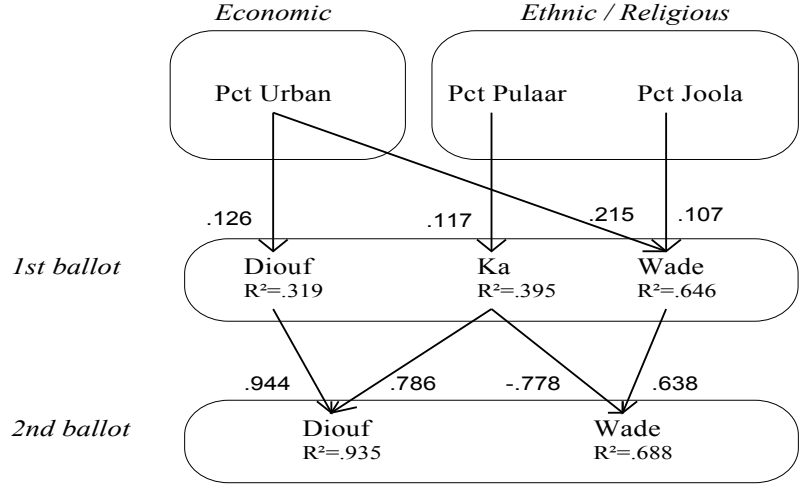

It is clear that the use of different *S*-degrees has allowed us to approach the data with more delicacy, and that the final model better reflects the complexity of the studied reality.

## 4.2. Dealing with interaction effects

So far, we only considered the predictors' marginal (constant) effects on the dependant variable. Let us now release this constraint by allowing a predictor's effect to be a linear function of other predictors' values.

### 4.2.1. Single equation model with interaction effects

#### 4.2.1.1. The model

Consider the following single equation model, in which a predictor interacts with some others:

$$F_s = \sum_{F_r \in J_s^t} \beta_r F_r + (\beta_t + \sum_{F_r \in J_t^s} \delta_{rt} F_r) F_t + \sum_{\substack{F_u \in E_s \setminus J_s^t \\ u \neq t}} \beta_u F_u$$

$F_s$ is the dependant variable. The set of its predictors is $E_s$. Amongst them, $F_t$ interacts with some other predictors, whose set is denoted $J_s^t$.

Supposing that all variables are currently known but $F_t$, we have to extend our internal estimation procedure so as to estimate $F_t$ (internal estimation will be denoted $\Phi_t$, as before).

Note: Should the model contain some interactions that do not involve $F_t$, each of the interacting predictors and of their products could be considered as part of the $F_u$'s in the last summation.

#### 4.2.1.2. Internal estimation procedure for interactive latent variables

Notation: $F$ being a variable and $X$ a variable group, $F \otimes X$ (or $X \otimes F$) will refer to the group formed by multiplying $F$ with every $x^j$ in $X$: $F \otimes X = \{ F x^j \mid x^j \in X \}$.

Algorithm:

*Initialization:*

$\Phi_t$'s initial value is calculated as if all $\delta_{rt}$ coefficients were zero, i.e. using the internal estimation scheme we proposed in section 4.1. for an explanatory factor.

*Current step 1:*

Regress $F_s$ onto $E_s \cup \{F_r \Phi_t\}_{F_r \in J_s^t}$. This provides with a current estimation of coefficients $\beta_t$ and $\delta_{rt}$, denoted: $\hat{\beta}_t, \hat{\delta}_{rt}$.

*Current step 2:*

1) Calculate group $Y_t = (\hat{\beta}_t + \sum_{F_r \in J_t^s} \hat{\delta}_{rt} F_r) \otimes X_t$ and use it as a new group in the internal estimation, to determine a factor $G_t$. This means regressing $F_s$ onto $Y_t$, $F_r$ and all non-interactive predictors, and extracting the $Y_t$-component $G_t$.

2) Then, divide $G_t$ by $(\hat{\beta}_t + \sum_{F_r \in J_t^s} \hat{\delta}_{rt} F_r)$ and standardize the result. This provides the new current value for $\Phi_t$. If this value is close enough from the previous one, stop. Else, go back to current step 1.

The illustration of this algorithm is given on figure 11 in a simplified case.

*Figure 11: Internal estimation of an interactive variable*

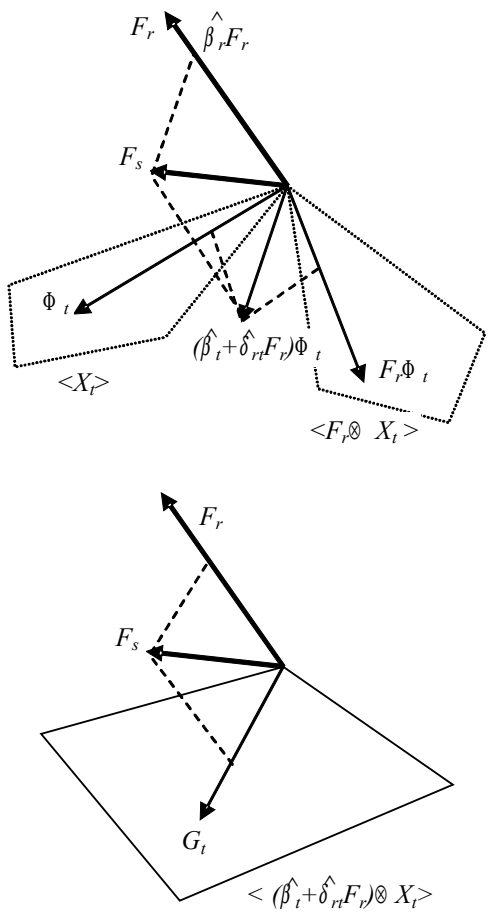

Dependant factor $F_s$ has only two predictors $F_r$ and $F_t$, which interact. We illustrate $F_t$'s internal estimation.

***Current step 1***

$F_s$ is regressed on $F_r, \Phi_t$ and $F_r\Phi_t$, which gives estimated coefficients $\hat{\beta}_r, \hat{\beta}_t, \hat{\delta}_{rt}$.

***Current step 2***

1) Subspace $\langle (\hat{\beta}_t + \hat{\delta}_{rt} F_r) \otimes X_t \rangle$ is formed.

Then, $F_s$ is regressed on this subspace and $F_r$, which gives a new component $G_t$.

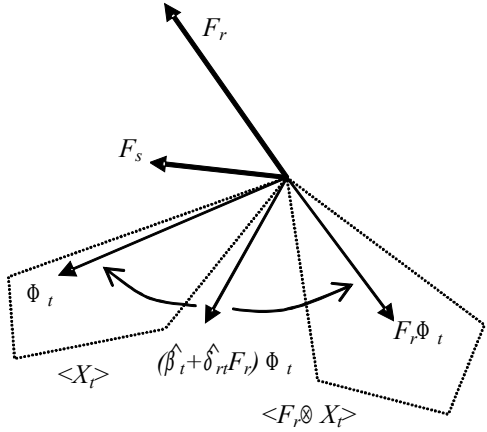

2) From equation $G_t = (\hat{\beta}_t + \hat{\delta}_{rt} F_r)\Phi_t$, we draw the new value of $\Phi_t$ (so, that of $F_r\Phi_t$).

Properties:

• R² increases:

Current step 1: The standard multiple regression optimizes R².

Current step 2: Subspace $< (\hat{\beta}_t + \sum_{F_r \in J_t^s} \hat{\delta}_{rt} F_r) \otimes X_t >$ contains the solution of step 1: $(\hat{\beta}_t + \sum_{F_r \in J_t^s} \hat{\delta}_{rt} F_r)\Phi_t$, but is larger. The regression performed here therefore increases R².

• When no $F_t$ in $X_t$ does really interact with any other predictor in $F_s$'s model, the procedure boils down to the formula used for models without interactions:

Current step 1 finds all $\hat{\delta}_{rt} = 0$. Therefore, subspace $< (\hat{\beta}_t + \sum_{F_r \in J_t^s} \hat{\delta}_{rt} F_r) \otimes X_t >$ formed in current step 2 is none other than $<X_t>$. So step 2 amounts to repeating initialization, i.e. what is done in models without interactions.

### 4.2.2. Multi-equation model with interaction effects

There is absolutely no difference with the technique used when no interaction is considered.

### 4.2.3. First application example: simulated data

• For 100 observations, we generated 20 independant random variables using uniform distribution $\cup[-1, 1]$, and aranged them in 2 groups, $A$ and $B$, containing 10 variables each: $a1 \ldots a10$ for group $A$, and $b1 \ldots b10$ for group $B$. Then, we computed the following $c$ variable:

$$c = 0.2(a1 + b1)/\sqrt{2/3} + 0.8(a2/10 + b2/10 + a2.b2)/\sqrt{53/450}$$

• All variables naturally have zero mean, including products $ai.bj$, since $E(ai.bj) = E(ai).E(bj) = 0$.

• Variable $c$ can be written: $c = 0.2\ c1 + 0.8\ c2$, where $c1 = (a1 + b1)/\sqrt{2/3}$ and $c2 = (a2/10 + b2/10 + a2*b2)/\sqrt{53/450}$ are two independant variables of unit variance:

$\forall i : V(ai) = V(bi) = (1+1)^2/12 = 1/3$ ; $\forall i,j : V(ai.bj) = E(ai.bj)^2 = E(ai^2).E(bj^2) = V(ai).V(bi) = 1/9$

$\forall i,j : \text{Cov}(ai,bj) = 0$ ; $\text{Cov}(ai, ai.bj) = E(ai \cdot ai.bj) = E(ai^2.bj) = E(ai^2).E(bj) = 0$

Similarly, $\text{Cov}(bj, ai.bj) = 0$.

As a consequence: V($a_1 + b_1$) = V($a_1$) + V($b_1$) = 2/3 and: V($a_2$/10 + $b_2$/10 + $a_2.b_2$) = V($a_2$)/100 + V($b_2$)/100 + V($a_2.b_2$) = 53/450. Therefore: V($c_1$) = V($c_2$) = 1.

- Note that coefficient of $c_1$ in $c$ is four times smaller than that of $c_2$. Variable $c_1$ contains no interaction between $a_1$ and $b_1$, whereas in $c_2$, interaction between $a_2$ and $b_2$ is dominating in terms of variance.

A method that only sees marginal effects should detect $a_1$ and $b_1$ as main components of $c$, whereas taking interactions into account should shift these components to $a_2$ and $b_2$ respectively.

- Notes: 1) Each of groups $A$ and $B$ consisting in uncorrelated variables having the same variance, it has no definite principal component system, so PCA prior to regression is no use at all.

2) The number of possible interactions between variables of groups $A$ and $B$ respectively amounts to 100 (it is the number of products $a_i.b_j$). If we add to this the number of marginal effects of both groups, we get 120 coefficients. As there are only 100 observations, it is impossible to regress $c$ onto $\{a_i, b_j, a_i.b_j\}_{ij}$ to estimate the model directly. So, the situation looks uncomfortable.

Let us submit the data to TCPM using linear resultant $S_0$ (it gives similar results to using no resultant at all, since groups have no definite PC structure), first without, then with interactions. Stability has been reached using 3 iterations inside the internal estimation procedure, and 15 iterations alternating internal and external estimations.

**TCPM without interactions:**

Here are the correlations between each factor we get and the variables of the corresponding group:

| Group $A$ | $a_1$ | $a_2$ | $a_3$ | $a_4$ | $a_5$ | $a_6$ | $a_7$ | $a_8$ | $a_9$ | $a_{10}$ |
|---|---|---|---|---|---|---|---|---|---|---|
| $FA$ | **.877** | .443 | -.760 | -.247 | .116 | .068 | .016 | .072 | -.143 | -.149 |

| Group $B$ | $b_1$ | $b_2$ | $b_3$ | $b_4$ | $b_5$ | $b_6$ | $b_7$ | $b_8$ | $b_9$ | $b_{10}$ |
|---|---|---|---|---|---|---|---|---|---|---|
| $FB$ | **.739** | .385 | -.216 | .206 | -.004 | .162 | .268 | .208 | -.230 | .103 |

TCPM without interactions, unable to detect the interaction between $a_2$ and $b_2$, tracked down the variables whose sole marginal effects were able to capture $c$'s variance best, i.e. $a_1$ and $b_1$ (factors are also positively, but weakly correlated to $a_2$ and $b_2$, respectively). But the part of explained variance remains modest (regressing $c$ onto $FA$ and $FB$ has $R^2$ = .307; regressing $c$ onto $a_1$ and $b_1$ has $R^2$ = .208). So, the procedure has missed the main phenomenon.

**TCPM with interactions:**

Correlations between each factor we get and the variables of the corresponding group are now:

| Group $A$ | $a_1$ | $a_2$ | $a_3$ | $a_4$ | $a_5$ | $a_6$ | $a_7$ | $a_8$ | $a_9$ | $a_{10}$ |
|---|---|---|---|---|---|---|---|---|---|---|
| $FA$ | .167 | **.969** | -.066 | .059 | -.177 | .002 | .115 | -.062 | .066 | -.369 |

| Group $B$ | $b_1$ | $b_2$ | $b_3$ | $b_4$ | $b_5$ | $b_6$ | $b_7$ | $b_8$ | $b_9$ | $b_{10}$ |
|---|---|---|---|---|---|---|---|---|---|---|
| $FB$ | -.020 | **-.979** | .110 | .172 | .026 | .144 | -.207 | .108 | .191 | -.078 |

Considering interactions allowed TCPM to detect the dominating variables in $c$'s model (regressing $c$ onto $FA$, $FB$ and $FAFB$ has now $R^2$ = .868; regressing $c$ onto $a_2$, $b_2$ and $a_2b_2$ has $R^2$ = .948).

*4.2.4. Second application example: modelling the rent in Dakar*

**The data (cf. appendix B):**

We are dealing with a sample of 41 houses let for rent in Dakar. For each one, we recorded the *monthly rent* (in thousands of FCFA, this single variable making up group X1, the dependant group), as well as three groups of explanatory characteristics:

House size characteristics (group X2):

- *Plot surface* (m²)  - *Built surface* (m²)

- *Built surface* / number of *residential rooms* (i.e. rooms except kitchens, bathrooms and WC)

- *Total number of rooms* (bedrooms, livingrooms, kitchens, bathrooms and WC)

- *Number of bathrooms*  - *Number of bedrooms*  - *Number of livingrooms*

- *Number of WC*  - *Number of kitchens*

Building quality characteristics (group X3):

- *Detached house* (0 = flat ; 1 = detached house)  - *Buiding standing* (0 = no ; 1 = yes)

- *General condition* (0 = poor ; 1 = medium ; 2 = fair ; 3 = new)  - *Garden* (0 = no ; 1 = yes)

- *Backyard* (0 = no ; 1 = yes)  - *Pool* (0 = no ; 1 = yes)  - *Garage* (0 = none ; 1 = single ; 2 = double)

- *High Tech* (number of high tech facilities, such as solar energy water heater, generating set, parabolic aerial...)

Area quality characteristics (group X4):

- *Distance to Town Centre* (0 = town center ; 1 = less than 2km from TC ; 2 = 2 - 10 km to TC ; 3 = over 10 km to TC)

- *Shopping area* (0 = more than 1km away ; 1 = less than 1km away)

- *Beach* (0 = more than 1km away ; 1 = less than 1km away)

- *Hotel businesses*, i.e. hotels, restaurants, casinos... (0 = more than 2km away ; 1 = less than 2km away)

- Access to one of the four *main roads* going to town centre (0 = more than 1km away ; 1 = less than 1km away)

- *Area standing* (0 = irregular ; 1 = lowerclass regular ; 2 = middleclass ; 3 = upperclass)

- *Business area* (0 = no ; 1 = yes)

The model that seemed natural to us is the following: *building quality* and *area quality* determine the *cost* of the house *per size unit* (*size* being a latent variable since it can be measured in various ways). Then, under the assumption that the return on investment is constant, *cost per size unit* and *size* should determine the *rent* in a multiplicative way (cf. fig 12).

*Figure 12: conceptual model of the rent*

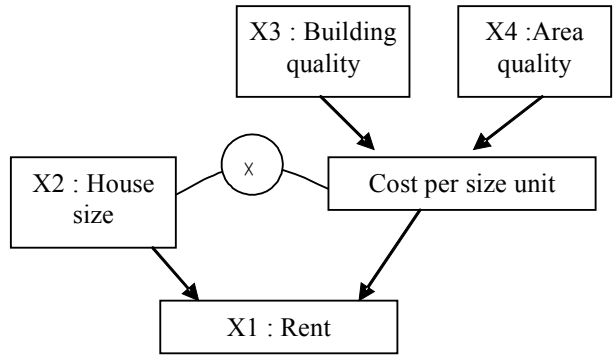

We shall use the following notations:

*Rent* = $R$ (observed) ; *Building quality* = $B$ (latent) ; *Area quality* = $A$ (latent) ; *Size* = $S$ (latent) ; *Cost per size unit* = $C$ (latent).

From the conceptual model shown on fig. 12, we draw the following equations:

$$\text{(a)} \quad R = r_0 + C\,S \quad ; \quad \text{(b)} \quad C = c_0 + c_A\,A + c_B\,B$$

Note that no observed variable can be taken as a measure of *Cost*. Therefore, we can not apply TCPM directly to our model. But removing *Cost* transforms this model so that every latent variable is directly supported by a group of observed variables. Indeed, from (a) and (b), we draw:

$$R = r_0 + (c_0 + c_A\,A + c_B\,B)S \quad \Leftrightarrow \quad \text{(c)} \quad R = r_0 + c_0\,S + c_A\,SA + c_B\,SB$$

We finally get a single equation model (c) involving two interactions. The corresponding conceptual model is shown on figure 13.

*Figure 13: alternative conceptual model of the rent*

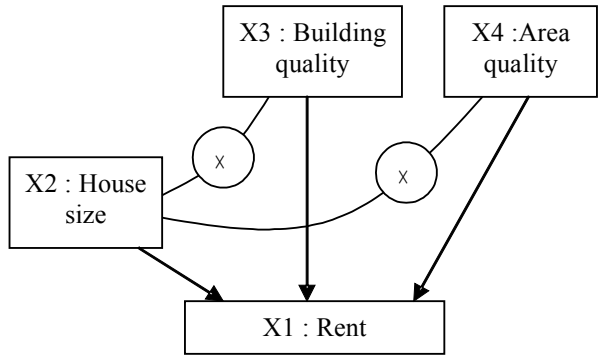

**Estimation :**

We computed several TCPM estimations:

- Without external estimation ; first without, then with interaction effects.

- With external estimation ; first without, then with interaction effects. We tried resultant orders 0 to 4, and found that $k = 1$ provided the best adjustment quality.

When estimation was made taking no account of interactions, the corresponding factor products were still calculated and put into the final regression model, to allow comparison.

The results of the estimations are compared in tables 3 to 6. Stability was always achieved in less than 10 iterations.

*Model adjustment quality:*

*Table 3 : Model adjustment ($R^2$)*

| Model<br>Estimation | S,B,A,SB,SA | S,B,A,SB | S,B,A,SA | S,B,A |
|---|---|---|---|---|
| Without external estimation | | | | |
| Without interactions | .934 | .921 | .933 | .917 |
| With interactions | **.993** | .885 | .977 | .830 |
| With $k = 0$ external estimation | | | | |
| Without interactions | .917 | .844 | .857 | .783 |
| With interactions | **.925** | .839 | .858 | .757 |
| With $k = 1$ external estimation | | | | |
| Without interactions | .922 | .846 | .868 | .795 |
| With interactions | **.945** | .845 | .843 | .756 |

The best adjusted model is, unsurprisingly, that estimated with interactions and without external estimation ($R^2$ is then the sole criterium to be maximized). The remarkable adjustment quality is due to several facts: 1) Pure $R^2$ maximization when one skips external estimations; 2) Taking interactions into account raises the number of predictive dimensions; 3) Rent fluctuations, given size and quality, are relatively small compared to global fluctuations, since houses go from timeworn single room to luxury detached house; 4) The number of explanatory observed variables is high, considering the number of observations. Note that when one skips external estimation, taking interactions into account improves adjustment quality (rising from 0.934 to 0.993) much more than when one does not. This is perfectly expectable, as external estimation restricts estimation liberty. Note also the adjustment quality loss when interaction terms *SB* and *SA* are being omitted in the model. Since $k = 1$ provides a better adjusted latent model, we shall retain this order for external estimation.

**Factor interpretation:**

A factor can be interpreted in two complementary ways:

1 - Using its correlations with the group's observed variables. In this case, the stress is put on the global relation between the factor and each of the observed variables.

2 - Using the coefficients of the observed variables in the factor's formula. In that case, the stress is put on the part each observed variable plays in the factor (partial relation). Here, we have standardized all variables so as to be able to compare their coefficients in absolute value.

*Table 4: Size factor*

| Correlations | Plot surface | Built surface | Built surf. / resid. room | Nr rooms | Nr résid. rooms | Nr bathrooms | Nr bedrooms | Nr livingrooms | Nr WC | Nr Kitchens |
|---|---|---|---|---|---|---|---|---|---|---|
| *Without external estimation* | | | | | | | | | | |
| No interactions | .915 | .766 | .658 | .710 | .627 | .698 | .584 | .620 | .823 | .264 |
| Interactions | .922 | .932 | .814 | .875 | .813 | .845 | .759 | .801 | .888 | .330 |
| *With external estimation* | | | | | | | | | | |
| No interactions | .854 | .985 | .773 | .977 | .945 | .913 | .911 | .858 | .902 | .331 |
| Interactions | .840 | .985 | .765 | .982 | .953 | .912 | .919 | .864 | .903 | .330 |
| **Coefficients** | | | | | | | | | | |
| *Without external estimation* | | | | | | | | | | |
| No interactions | .836 | .634 | -.500 | -.083 | -.336 | -.078 | -.464 | .030 | .778 | .053 |
| Interactions | .351 | .980 | -.228 | -.072 | -.243 | -.074 | -.363 | .090 | .366 | -.055 |
| *With external estimation* | | | | | | | | | | |
| No interactions | .172 | .159 | .084 | .118 | .107 | .118 | .097 | .111 | .135 | .006 |
| Interactions | .132 | .168 | .083 | .124 | .117 | .113 | .109 | .116 | .134 | .006 |

**House size** (cf. table 4):

The estimated *size* factor is strongly and positively correlated with all size variables, which entitles us to interpret it as such in all cases. It is generaly closer to the *built surface* than to any other variable. The least relevant variable appears to be the number of kitchens (a nearly constant variable, since in the great majority of houses, there is a single kitchen, with the exception of isolate rooms, having none, and few luxury detached houses having two).

As for coefficients, they show a great disparity across estimations. When one skips external estimation, signs and absolute values vary considerably: effect transfers occur between correlated variables, obviously. The role of external estimation is to shrink such transfers, and so so it does, giving only positive coefficients having balanced absolute values. The only variable having a coefficient much weaker than the others' is the number of kitchens, already noted as little relevant.

*Table 5: Building quality factor*

| **Correlations** | Detached house | Standing | Condition | Garden | Back-yard | Pool | Garage | High Tech |
|---|---|---|---|---|---|---|---|---|
| *Without external estimation* | | | | | | | | |
| No interactions | .648 | .023 | .421 | -.157 | .058 | -.067 | .613 | .431 |
| Interactions | .840 | .541 | .631 | .445 | .578 | .313 | .902 | .591 |
| *With external estimation* | | | | | | | | |
| No interactions | .419 | .764 | .643 | .644 | .222 | .411 | .694 | .885 |
| Interactions | .475 | .849 | .756 | .669 | .276 | .434 | .744 | .694 |
| **Coefficients** | | | | | | | | |
| *Without external estimation* | | | | | | | | |
| No interactions | .653 | -.187 | .430 | -.602 | -.384 | -.270 | .200 | .435 |
| Interactions | .341 | .095 | .143 | .058 | .282 | -.102 | .389 | .108 |
| *With external estimation* | | | | | | | | |
| No interactions | -.000 | .366 | .196 | .076 | -.003 | .016 | .028 | .588 |
| Interactions | .002 | .432 | .357 | .171 | .001 | .028 | .151 | .178 |

**Building quality** (cf. table 5):

Skipping external estimation and considering no interaction lead to a factor poorly correlated with quality variables and with unconstant sign. Coefficients also have heterogenous absolute values and signs. Under such circumstances, factor interpretation is awkward. Taking interactions into account improves the situation a great deal: the factor it provides is positively and often well correlated with the variables that mean quality rise. Moreover, when one uses external estimation, coefficients of all variables in the factor's formula become all positive.

*Table 6 : Area quality factor*

| Correlations | Distance to TC | Shopping area | Beach | Hotel businesses | Main road | Area standing | Business area |
|---|---|---|---|---|---|---|---|
| *Without external estimation* | | | | | | | |
| No interactions | -.432 | .565 | .174 | .608 | .517 | .492 | .571 |
| Interactions | -.637 | .270 | .027 | .572 | .539 | .388 | .887 |
| *With external estimation* | | | | | | | |
| No interactions | -.919 | .493 | -.360 | .539 | .204 | .179 | .853 |
| Interactions | -.860 | .328 | -.324 | .462 | .266 | .221 | .941 |
| **Coefficients** | | | | | | | |
| *Without external estimation* | | | | | | | |
| No interactions | .212 | .704 | .202 | .227 | .250 | .258 | .462 |
| Interactions | -.115 | .180 | .219 | .160 | .249 | .012 | .723 |
| *With external estimation* | | | | | | | |
| No interactions | -.455 | .172 | -.000 | .181 | .035 | .022 | .456 |
| Interactions | -.366 | .032 | -.001 | .112 | .051 | .018 | .643 |

**Area quality** (cf. table 6):

Here, all variables *a priori* mean a facility, with the possible exception of *Distance to TC*. Estimations provide a factor negatively correlated to the latter variable, and positively to *Business area* (the business area being located in the town centre). Skipping external estimation and ignoring interactions leads to a factor poorly correlated with the group's variables, and with coefficients sometimes having irrelevant signs (e.g. *Distance to TC*'s and *Business area*'s have the same sign). Note that merely introducing interactions yields results much easier to interpret. Besides, what external estimation does is very clear: it draws the latent variable towards the Town Center (business area) factor, which seems to be the most important in this group.

**Conclusion :**

Taking interactions into account always provided a better fitting model. Besides, external estimation proved essential for factor interpretation. Therefore, we will retain the model estimated through the last procedure. This estimated model (linking standardized variables) is:

$$R = -.123 + .584\ S + .280\ B + .598\ A + .318\ SB + .542\ SA$$

# 5. Conclusion

In these developments, we have kept the basic idea of PLSPM, i.e. alternating latent model adjustment (internal estimation) and attraction of factors to strong correlation structures in groups (external estimation). But both internal and external estimation mechanisms have been extended, so as to be able to focus on a greater variety of phenomenons: interactive latent variables in the latent model, and the existence of bundles in groups. As the application examples show, the main asset of these extensions is that they give a greater flexibility to the Path Modelling technique, and allow to better respect the complexity of things during exploration. Of course, the analyst will have to pay for it, by trying several options as to the model (the choice of interactions) and the external estimation tool (the resultant's order). Then, he/she will have to examine the results produced by these choices, and take the responsability to pick some and discard others. But we feel that there lies perhaps the second main asset of the extensions: when selection is possible, it has to be supported by valuable arguments, so the analyst may have to think things over with greater care.

**Thanks**

Very warm thanks to Pierre Cazes for careful reading and clever advising.

# Appendices :

## A - Senegalese Departmental Data[2]

| District | NHPI | PctAgriInc | IncActivePers | ActivePop | Scol | Malnutrition | DrinkWater | Rural | Urban | PopDensity | HouseholdSize | Pop0_14 | Pop15_60 | PopOver60 | WIndep | WPublic | Wprivate | WApprentice |
|---|---|---|---|---|---|---|---|---|---|---|---|---|---|---|---|---|---|---|
| BAKEL | 59.5 | 12.4 | 692 | 55.7 | 19.1 | 0 | 52.5 | 84.7 | 15.3 | 6 | 8.9 | 45.8 | 50.0 | 4.3 | 56.6 | 2.0 | 7.0 | 34.4 |
| BAMBEY | 64.1 | 20.6 | 213 | 53.7 | 19.9 | 5.9 | 27.6 | 90.3 | 9.7 | 160 | 9.5 | 49.3 | 43.5 | 7.1 | 58.7 | 0.4 | 2.7 | 38.2 |
| BIGNONA | 53.9 | 35.1 | 238 | 54.1 | 62.6 | 5.3 | 13.1 | 66.1 | 33.9 | 37 | 8.2 | 49.5 | 44.3 | 6.2 | 60.3 | 2.4 | 2.8 | 34.5 |
| DAGANA | 38.5 | 26.3 | 350 | 25.2 | 46.0 | 11.3 | 66.3 | 39.9 | 60.1 | 276 | 10.0 | 46.0 | 49.2 | 4.8 | 42.5 | 6.0 | 14.7 | 36.7 |
| DAKAR | 1.5 | 0 | 9722 | 57.0 | 76.5 | 6 | 97.9 | 0.0 | 100.0 | 28027 | 7.0 | 38.1 | 58.6 | 3.2 | 33.5 | 13.8 | 38.2 | 14.5 |
| DIOURBEL | 57.5 | 24.1 | 202 | 34.0 | 35.3 | 2.8 | 50 | 52.5 | 47.5 | 170 | 8.9 | 49.2 | 45.8 | 5.0 | 58.3 | 1.4 | 6.3 | 34.0 |
| FATICK | 75.6 | 23.1 | 211 | 62.0 | 41.6 | 10 | 21.1 | 90.2 | 9.8 | 84 | 8.2 | 52.5 | 42.1 | 5.4 | 50.6 | 0.8 | 1.4 | 47.2 |
| FOUNDIOUGNE | 52.9 | 46 | 249 | 58.0 | 17.3 | 3.3 | 11 | 90.8 | 9.2 | 142 | 10.3 | 52.6 | 42.1 | 5.3 | 64.9 | 0.4 | 1.6 | 33.1 |
| GOSSAS | 63.0 | 28.4 | 317 | 52.7 | 7.5 | 5.4 | 63.1 | 87.8 | 12.2 | 116 | 9.4 | 48.9 | 44.9 | 6.2 | 66.3 | 0.3 | 1.6 | 31.9 |
| KAFFRINE | 66.4 | 35.5 | 206 | 70.3 | 8.9 | 5 | 32.9 | 94.3 | 5.7 | 55 | 9.4 | 49.4 | 46.2 | 4.4 | 63.4 | 0.7 | 0.4 | 35.5 |
| KAOLACK | 44.4 | 42.5 | 996 | 27.6 | 35.9 | 4.2 | 62.9 | 49.0 | 51.0 | 227 | 9.0 | 47.2 | 48.0 | 4.8 | 59.0 | 3.4 | 5.6 | 32.0 |
| KEBEMER | 61.2 | 14 | 658 | 58.9 | 16.7 | 4.9 | 76 | 87.5 | 12.5 | 43 | 9.2 | 47.7 | 46.6 | 5.7 | 58.4 | 0.5 | 1.2 | 39.8 |
| KEDOUGOU | 84.1 | 33.7 | 184 | 53.6 | 18.5 | 4.9 | 21.5 | 100.0 | 0.0 | 4 | 8.5 | 47.7 | 47.7 | 4.5 | 60.0 | 0.2 | 0.3 | 39.5 |
| KOLDA | 71.7 | 52.6 | 349 | 45.8 | 29.8 | 5.8 | 2.8 | 77.2 | 22.8 | 25 | 8.1 | 49.2 | 46.9 | 3.9 | 54.3 | 1.7 | 1.2 | 42.8 |
| LINGUERE | 72.0 | 9.8 | 1150 | 55.7 | 18.0 | 15 | 53 | 92.1 | 7.9 | 8 | 7.9 | 46.7 | 47.2 | 6.1 | 56.6 | 0.3 | 1.0 | 42.1 |
| LOUGA | 48.0 | 13.3 | 574 | 42.0 | 18.8 | 16.2 | 55.3 | 75.3 | 24.7 | 36 | 9.0 | 48.1 | 46.9 | 5.0 | 55.6 | 2.0 | 2.7 | 39.6 |
| MATAM | 71.9 | 2.8 | 703 | 56.9 | 9.2 | 6 | 18.8 | 100.0 | 0.0 | 10 | 9.0 | 49.9 | 43.9 | 6.2 | 43.7 | 0.1 | 1.3 | 54.9 |
| MBACKE | 65.4 | 6.5 | 102 | 47.9 | 7.1 | 1.2 | 76.8 | 84.9 | 15.1 | 407 | 8.3 | 48.1 | 45.3 | 6.6 | 73.8 | 0.6 | 5.7 | 19.9 |
| MBOUR | 55.8 | 7.4 | 199 | 36.7 | 41.7 | 0 | 49 | 68.2 | 31.8 | 506 | 9.4 | 51.0 | 43.6 | 5.4 | 60.2 | 2.2 | 13.1 | 24.5 |
| NIORO | 62.0 | 56.5 | 627 | 52.5 | 17.1 | 4.3 | 19.1 | 92.2 | 7.8 | 91 | 9.7 | 51.6 | 44.6 | 3.8 | 62.3 | 1.3 | 0.7 | 35.6 |
| OUSSOUYE | 74.5 | 23.8 | 249 | 56.1 | 77.2 | 0 | 0 | 100.0 | 0.0 | 46 | 5.4 | 47.5 | 40.1 | 12.4 | 77.0 | 0.0 | 0.0 | 23.0 |
| PIKINE | 22.9 | 0 | 2778 | 51.0 | 55.7 | 6.1 | 89.9 | 0.0 | 100.0 | 21030 | 9.1 | 46.8 | 49.9 | 3.3 | 46.0 | 7.2 | 25.0 | 21.9 |
| PODOR | 63.3 | 8.2 | 396 | 59.4 | 32.6 | 18.2 | 29 | 88.4 | 11.6 | 12 | 8.0 | 50.0 | 44.4 | 5.6 | 48.4 | 3.0 | 10.7 | 38.0 |
| RUFISQUE | 15.7 | 10 | 1389 | 49.0 | 64.5 | 6 | 99.5 | 27.0 | 73.0 | 645 | 9.2 | 45.1 | 49.9 | 5.0 | 46.5 | 7.1 | 24.8 | 21.6 |
| SEDHIOU | 70.0 | 31.8 | 200 | 54.8 | 27.7 | 2.2 | 13.7 | 90.0 | 10.0 | 42 | 9.4 | 50.0 | 43.9 | 6.1 | 54.8 | 1.0 | 0.8 | 43.4 |
| TAMBACOUNDA | 68.1 | 26.5 | 843 | 44.7 | 22.3 | 8 | 18.5 | 81.1 | 18.9 | 11 | 8.3 | 49.3 | 46.9 | 3.8 | 61.5 | 1.6 | 2.9 | 34.0 |
| THIES | 37.8 | 25.6 | 144 | 30.3 | 42.3 | 3.1 | 66.9 | 50.6 | 49.4 | 464 | 10.2 | 48.6 | 46.2 | 5.3 | 47.4 | 5.5 | 9.4 | 37.7 |
| TIVAOUANE | 53.3 | 16.4 | 123 | 51.3 | 22.1 | 3.2 | 31.4 | 79.8 | 20.2 | 198 | 9.1 | 49.8 | 46.0 | 4.2 | 62.3 | 0.4 | 3.7 | 33.6 |
| VELINGARA | 65.8 | 35.1 | 591 | 25.6 | 21.6 | 2.8 | 0 | 84.5 | 15.5 | 26 | 7.3 | 48.5 | 46.9 | 4.6 | 68.5 | 1.0 | 0.6 | 29.9 |
| ZIGUINCHOR | 40.5 | 26.4 | 157 | 16.5 | 57.3 | 4.8 | 18.5 | 29.8 | 70.2 | 129 | 8.1 | 48.1 | 48.0 | 3.9 | 60.7 | 2.5 | 15.6 | 21.2 |

---

[2] Source : Direction de la Prévision et de la Statistique du Sénégal

| District | OEmployed | OUnemployed | OStudent | OHouseWife | ORetired | ASPrim | ASSec | ASTer | Wolof | Sereer | Joola | Pulaar | Manding | PctMoslims |
|---|---|---|---|---|---|---|---|---|---|---|---|---|---|---|
| BAKEL | 68.0 | 2.3 | 6.7 | 18.0 | 5.0 | 82.8 | 3.6 | 13.6 | 3.8 | 0.3 | 0.0 | 50.0 | 9.6 | 100.0 |
| BAMBEY | 76.6 | 1.7 | 8.0 | 8.3 | 5.4 | 86.2 | 2.7 | 11.1 | 57.3 | 38.9 | 0.1 | 2.9 | 0.1 | 99.1 |
| BIGNONA | 51.1 | 3.6 | 31.5 | 7.4 | 6.4 | 78.2 | 3.4 | 18.4 | 1.8 | 1.2 | 80.6 | 5.2 | 6.1 | 77.0 |
| DAGANA | 43.0 | 10.0 | 15.8 | 26.6 | 4.6 | 53.0 | 14.8 | 32.2 | 63.6 | 1.3 | 0.7 | 31.2 | 1.4 | 98.6 |
| DAKAR | 37.8 | 16.1 | 24.3 | 17.3 | 4.5 | 2.3 | 15.3 | 82.3 | 49.1 | 13.0 | 6.9 | 16.5 | 5.6 | 87.2 |
| DIOURBEL | 61.6 | 4.1 | 14.4 | 14.1 | 5.9 | 63.1 | 9.6 | 27.3 | 53.4 | 34.4 | 0.4 | 9.4 | 0.5 | 100.0 |
| FATICK | 71.3 | 1.9 | 15.5 | 6.8 | 4.6 | 93.6 | 1.4 | 4.9 | 29.9 | 86.0 | 0.1 | 5.1 | 1.3 | 78.9 |
| FOUNDIOUGNE | 78.6 | 2.2 | 6.9 | 7.2 | 5.2 | 89.3 | 1.4 | 9.3 | 6.1 | 37.7 | 0.6 | 9.0 | 9.3 | 98.8 |
| GOSSAS | 83.0 | 1.1 | 1.6 | 10.3 | 4.1 | 93.1 | 0.8 | 6.1 | 39.1 | 29.8 | 0.1 | 14.6 | 1.1 | 100.0 |
| KAFFRINE | 79.4 | 1.3 | 5.6 | 8.2 | 5.5 | 93.2 | 1.3 | 5.5 | 71.5 | 6.0 | 0.2 | 18.7 | 2.3 | 98.4 |
| KAOLACK | 56.1 | 4.4 | 16.7 | 18.0 | 4.7 | 61.3 | 7.1 | 31.6 | 47.3 | 22.8 | 0.5 | 19.7 | 5.3 | 99.3 |
| KEBEMER | 75.7 | 6.5 | 4.9 | 9.7 | 3.2 | 79.6 | 2.3 | 18.1 | 82.6 | 1.0 | 0.0 | 17.0 | 0.0 | 100.0 |
| KEDOUGOU | 84.2 | 0.9 | 2.2 | 6.8 | 5.9 | 98.3 | 0.2 | 1.4 | 1.4 | 0.4 | 0.0 | 41.0 | 35.0 | 93.3 |
| KOLDA | 75.9 | 2.2 | 10.3 | 4.9 | 6.7 | 87.3 | 3.3 | 9.4 | 7.6 | 0.2 | 1.6 | 73.5 | 9.7 | 96.5 |
| LINGUERE | 73.7 | 3.6 | 5.3 | 11.9 | 5.5 | 87.0 | 1.5 | 11.6 | 46.9 | 4.5 | 0.5 | 48.0 | 0.5 | 99.4 |
| LOUGA | 72.0 | 9.9 | 6.7 | 8.4 | 3.0 | 77.7 | 3.1 | 19.2 | 75.8 | 0.5 | 0.1 | 23.0 | 0.1 | 99.3 |
| MATAM | 62.7 | 2.3 | 4.9 | 24.5 | 5.6 | 90.3 | 1.7 | 8.0 | 3.9 | 0.1 | 0.0 | 88.8 | 0.3 | 100.0 |
| MBACKE | 66.7 | 3.1 | 6.7 | 16.8 | 6.7 | 39.8 | 6.4 | 53.8 | 84.9 | 5.5 | 0.1 | 8.4 | 0.1 | 100.0 |
| MBOUR | 54.2 | 6.9 | 14.0 | 16.7 | 8.2 | 56.8 | 5.7 | 37.5 | 26.9 | 57.6 | 0.8 | 10.8 | 2.9 | 92.5 |
| NIORO | 70.3 | 5.5 | 10.2 | 6.8 | 7.1 | 91.0 | 1.1 | 7.9 | 70.7 | 4.1 | 0.0 | 21.4 | 2.0 | 99.5 |
| OUSSOUYE | 57.2 | 1.3 | 32.3 | 2.6 | 6.5 | 88.3 | 7.0 | 4.7 | 4.8 | 3.5 | 82.4 | 4.7 | 1.5 | 25.2 |
| PIKINE | 36.3 | 13.5 | 19.7 | 26.6 | 3.9 | 3.2 | 21.5 | 75.4 | 3.5 | 10.6 | 3.5 | 22.3 | 4.2 | 95.6 |
| PODOR | 38.8 | 6.5 | 14.6 | 33.1 | 6.9 | 63.4 | 6.1 | 30.5 | 5.5 | 0.3 | 0.1 | 92.1 | 0.2 | 100.0 |
| RUFISQUE | 36.2 | 11.1 | 22.0 | 25.4 | 5.4 | 18.1 | 19.5 | 62.4 | 1.3 | 9.8 | 1.3 | 12.8 | 2.6 | 95.6 |
| SEDHIOU | 78.8 | 0.8 | 11.4 | 2.7 | 6.3 | 93.2 | 1.7 | 5.1 | 1.6 | 0.2 | 10.9 | 19.9 | 39.5 | 82.3 |
| TAMBACOUNDA | 65.5 | 2.8 | 7.3 | 20.6 | 3.8 | 80.4 | 3.0 | 16.6 | 14.4 | 5.6 | 0.6 | 43.6 | 21.7 | 98.6 |
| THIES | 44.0 | 10.9 | 15.9 | 20.7 | 8.5 | 47.8 | 7.4 | 44.9 | 53.7 | 26.9 | 1.0 | 13.7 | 3.1 | 97.3 |
| TIVAOUANE | 65.4 | 9.1 | 9.6 | 11.9 | 3.9 | 72.5 | 4.8 | 22.7 | 80.1 | 8.1 | 0.4 | 10.0 | 0.8 | 100.0 |
| VELINGARA | 75.3 | 2.4 | 5.9 | 11.5 | 4.9 | 87.2 | 3.4 | 9.4 | 1.2 | 0.1 | 0.7 | 80.0 | 8.3 | 96.9 |
| ZIGUINCHOR | 48.7 | 10.1 | 25.3 | 11.2 | 4.7 | 44.3 | 12.4 | 43.3 | 8.2 | 3.4 | 34.4 | 13.5 | 14.4 | 67.1 |

| Votes for (%): District | Thiam1 | Niasse1 | Ka1 | Wade1 | Dieye1 | Sock1 | Fall1 | Diouf1 | abstention1 | Diouf2 | Wade2 | abstention2 |
|---|---|---|---|---|---|---|---|---|---|---|---|---|
| BAKEL | 0.6 | 3.6 | 3.7 | 14.4 | 0.6 | 0.4 | 0.4 | 33.2 | 43.0 | 32.0 | 25.9 | 42.1 |
| BAMBEY | 1.2 | 5.9 | 1.6 | 19.1 | 0.6 | 0.4 | 2.1 | 30.6 | 38.6 | 23.8 | 39.8 | 36.4 |
| BIGNONA | 0.9 | 5.5 | 1.3 | 28.5 | 0.4 | 0.5 | 0.3 | 23.6 | 38.9 | 20.9 | 41.4 | 37.7 |
| DAGANA | 0.8 | 4.4 | 9.5 | 18.5 | 0.9 | 0.3 | 0.3 | 36.7 | 28.7 | 40.2 | 29.0 | 30.8 |
| DAKAR | 0.7 | 16.3 | 3.0 | 33.8 | 0.4 | 0.2 | 0.7 | 15.2 | 29.6 | 15.5 | 50.1 | 34.4 |
| DIOURBEL | 1.3 | 10.5 | 3.3 | 16.5 | 0.7 | 0.5 | 2.4 | 28.0 | 36.8 | 24.7 | 39.2 | 36.1 |
| FATICK | 1.2 | 14.8 | 2.2 | 9.6 | 0.7 | 0.5 | 0.6 | 38.1 | 32.4 | 35.4 | 33.0 | 31.6 |
| FOUNDIOUGNE | 0.8 | 22.3 | 1.8 | 7.7 | 0.4 | 0.3 | 0.5 | 29.3 | 36.8 | 29.5 | 36.0 | 34.4 |
| GOSSAS | 0.9 | 9.7 | 4.0 | 13.1 | 0.7 | 0.5 | 1.1 | 30.9 | 39.1 | 29.5 | 34.0 | 36.5 |
| KAFFRINE | 0.8 | 10.5 | 2.9 | 7.1 | 0.6 | 0.4 | 0.6 | 29.4 | 47.8 | 31.5 | 24.9 | 43.6 |
| KAOLACK | 0.8 | 24.3 | 4.2 | 11.4 | 0.6 | 0.4 | 0.7 | 23.6 | 33.9 | 23.6 | 42.2 | 34.2 |
| KEBEMER | 0.7 | 3.4 | 2.7 | 15.0 | 0.3 | 0.2 | 0.5 | 24.9 | 52.4 | 23.3 | 26.5 | 50.2 |
| KEDOUGOU | 1.0 | 3.3 | 7.4 | 11.9 | 2.5 | 1.0 | 0.5 | 28.0 | 44.4 | 24.0 | 31.6 | 44.4 |
| KOLDA | 0.7 | 5.9 | 4.5 | 23.4 | 1.0 | 0.7 | 0.4 | 20.6 | 42.8 | 21.3 | 40.5 | 38.3 |
| LINGUERE | 0.4 | 1.7 | 23.2 | 5.9 | 0.8 | 0.3 | 0.3 | 31.3 | 36.0 | 45.8 | 15.4 | 38.8 |
| LOUGA | 0.7 | 6.4 | 4.8 | 10.1 | 0.7 | 0.6 | 0.7 | 39.0 | 37.0 | 40.0 | 21.6 | 38.3 |
| MATAM | 0.4 | 3.6 | 12.0 | 4.6 | 0.7 | 0.3 | 0.3 | 33.3 | 45.0 | 37.8 | 15.3 | 46.9 |
| MBACKE | 0.5 | 5.4 | 2.1 | 16.3 | 0.3 | 0.2 | 1.4 | 22.9 | 50.9 | 18.0 | 31.6 | 50.5 |
| MBOUR | 0.7 | 10.5 | 2.6 | 14.4 | 0.5 | 0.3 | 0.5 | 36.4 | 34.0 | 34.2 | 30.9 | 34.9 |
| NIORO | 0.7 | 29.9 | 2.0 | 6.0 | 0.6 | 0.3 | 0.5 | 28.7 | 31.3 | 29.7 | 42.2 | 28.1 |
| OUSSOUYE | 0.8 | 7.2 | 1.7 | 16.4 | 0.3 | 0.4 | 0.4 | 29.9 | 42.8 | 26.2 | 31.0 | 42.8 |
| PIKINE | 0.6 | 11.4 | 4.2 | 30.4 | 0.3 | 0.2 | 0.7 | 12.5 | 39.6 | 13.2 | 43.9 | 43.0 |
| PODOR | 0.4 | 4.1 | 16.4 | 5.4 | 0.7 | 0.3 | 0.2 | 32.1 | 40.4 | 41.5 | 16.6 | 41.8 |
| RUFISQUE | 0.9 | 10.8 | 3.1 | 31.1 | 0.7 | 0.3 | 0.6 | 30.7 | 21.7 | 26.0 | 49.9 | 24.1 |
| SEDHIOU | 1.2 | 9.5 | 1.7 | 16.5 | 0.5 | 0.4 | 0.6 | 27.7 | 41.9 | 24.8 | 35.5 | 39.7 |
| TAMBACOUNDA | 1.0 | 7.3 | 5.3 | 13.7 | 1.4 | 0.7 | 0.7 | 30.8 | 39.1 | 31.5 | 29.5 | 39.0 |
| THIES | 0.6 | 6.8 | 1.8 | 25.3 | 0.4 | 0.2 | 1.0 | 20.9 | 42.9 | 19.7 | 37.8 | 42.5 |
| TIVAOUANE | 0.8 | 5.2 | 1.9 | 23.2 | 0.7 | 0.2 | 0.8 | 32.2 | 34.8 | 30.1 | 36.9 | 32.9 |
| VELINGARA | 0.9 | 7.3 | 4.0 | 15.2 | 0.8 | 0.5 | 0.4 | 32.2 | 38.8 | 29.4 | 32.7 | 37.9 |
| ZIGUINCHOR | 0.8 | 11.5 | 3.4 | 22.7 | 0.5 | 0.5 | 0.3 | 21.3 | 38.9 | 19.4 | 40.8 | 39.9 |

***B - 41 houses in Dakar***

| Quartier | Monthly Rent | Plot Surface | Built Surface | Built Surf/Resid room | Nr Rooms | Nr Resid Rooms | Nr Bathrooms | Nr Bedrooms | Nt Livingrooms | Nr WC | Nr Kitchens | Detached House | House standing | Condition | Garden | Backyard | Pool | Garage | High Tech | Dist to TC | Shpping Area | Beach | Hotel buis. | Main Road | Area Standing | Business Area |
|---|---|---|---|---|---|---|---|---|---|---|---|---|---|---|---|---|---|---|---|---|---|---|---|---|---|---|
| Fass | 20 | 19 | 19 | 19 | 3 | 1 | 1 | 1 | 0 | 1 | 0 | 0 | 0 | 1 | 0 | 0 | 0 | 0 | 0 | 1 | 1 | 0 | 1 | 0 | 1 | 0 |
| Fass | 30 | 55 | 55 | 27.5 | 5 | 2 | 1 | 1 | 1 | 1 | 1 | 0 | 1 | 0 | 0 | 0 | 0 | 0 | 2 | 1 | 0 | 1 | 0 | 0 | 0 |
| Derkle | 35 | 40 | 40 | 13.3 | 6 | 3 | 1 | 2 | 1 | 1 | 1 | 0 | 1 | 0 | 0 | 0 | 0 | 0 | 2 | 1 | 0 | 0 | 0 | 2 | 0 |
| Colobane | 48 | 58 | 58 | 19.3 | 7 | 3 | 1 | 2 | 1 | 2 | 1 | 0 | 0 | 0 | 1 | 0 | 0 | 0 | 2 | 1 | 0 | 0 | 1 | 1 | 0 |
| Bopp | 50 | 75 | 75 | 15.0 | 9 | 5 | 1 | 3 | 2 | 2 | 1 | 0 | 0 | 2 | 0 | 1 | 0 | 0 | 0 | 2 | 1 | 0 | 0 | 0 | 1 | 0 |
| GrandYoff | 50 | 80 | 80 | 20.0 | 9 | 4 | 2 | 3 | 1 | 2 | 1 | 0 | 0 | 1 | 0 | 1 | 0 | 1 | 0 | 3 | 0 | 0 | 0 | 1 | 1 | 0 |
| LiberteI | 50 | 36 | 36 | 18.0 | 5 | 2 | 1 | 1 | 1 | 1 | 1 | 0 | 1 | 3 | 0 | 1 | 0 | 0 | 0 | 2 | 1 | 0 | 0 | 0 | 2 | 0 |
| Malika | 55 | 99 | 129 | 25.8 | 9 | 5 | 1 | 3 | 2 | 2 | 1 | 1 | 0 | 1 | 0 | 1 | 0 | 0 | 0 | 3 | 0 | 1 | 0 | 0 | 1 | 0 |
| Yoff | 55 | 53 | 53 | 17.7 | 6 | 3 | 1 | 2 | 1 | 1 | 1 | 0 | 0 | 2 | 0 | 1 | 0 | 0 | 0 | 3 | 1 | 1 | 0 | 1 | 1 | 0 |
| NiayeCoker | 60 | 59 | 59 | 14.8 | 7 | 4 | 1 | 3 | 1 | 1 | 1 | 0 | 0 | 1 | 0 | 1 | 0 | 0 | 0 | 1 | 1 | 0 | 1 | 0 | 1 | 0 |
| Pikine | 60 | 60 | 60 | 15 | 7 | 4 | 1 | 3 | 1 | 1 | 1 | 0 | 0 | 2 | 0 | 1 | 0 | 0 | 0 | 3 | 1 | 0 | 0 | 0 | 1 | 0 |
| JetdEau | 70 | 69 | 69 | 17.3 | 8 | 4 | 1 | 3 | 1 | 2 | 1 | 0 | 1 | 2 | 1 | 1 | 0 | 1 | 1 | 2 | 1 | 0 | 0 | 0 | 2 | 0 |
| Medina | 90 | 70 | 70 | 17.5 | 9 | 4 | 2 | 3 | 1 | 2 | 1 | 0 | 0 | 1 | 0 | 1 | 0 | 0 | 0 | 1 | 1 | 0 | 1 | 1 | 1 | 0 |
| Medina | 90 | 50 | 50 | 16.7 | 6 | 3 | 1 | 2 | 1 | 1 | 0 | 0 | 2 | 0 | 1 | 0 | 0 | 0 | 1 | 1 | 0 | 1 | 1 | 1 | 1 |
| Yoff | 90 | 71 | 71 | 23.7 | 6 | 3 | 1 | 2 | 1 | 1 | 1 | 0 | 1 | 0 | 1 | 0 | 0 | 0 | 3 | 1 | 1 | 1 | 1 | 1 | 0 |
| Castors | 95 | 154 | 139 | 19.9 | 11 | 7 | 1 | 6 | 1 | 2 | 1 | 1 | 0 | 0 | 0 | 1 | 0 | 0 | 0 | 2 | 1 | 0 | 0 | 0 | 2 | 0 |
| GueuleTapee | 95 | 90 | 90 | 18 | 10 | 5 | 2 | 4 | 1 | 2 | 1 | 0 | 0 | 1 | 0 | 1 | 0 | 0 | 0 | 1 | 1 | 0 | 1 | 1 | 1 | 0 |
| HLM | 120 | 149 | 149 | 21.3 | 12 | 7 | 2 | 5 | 2 | 2 | 1 | 1 | 0 | 1 | 0 | 1 | 0 | 2 | 0 | 2 | 1 | 0 | 0 | 0 | 1 | 0 |
| LiberteVI | 145 | 147 | 176 | 22.0 | 14 | 8 | 3 | 6 | 2 | 2 | 1 | 1 | 0 | 2 | 0 | 1 | 0 | 2 | 0 | 2 | 1 | 0 | 0 | 0 | 2 | 0 |
| SacreCoeurIII | 150 | 205 | 159 | 26.5 | 11 | 6 | 2 | 4 | 2 | 2 | 1 | 1 | 0 | 3 | 0 | 1 | 0 | 2 | 0 | 2 | 0 | 0 | 0 | 1 | 2 | 0 |
| Parcelles | 190 | 148 | 292 | 22.5 | 20 | 13 | 3 | 10 | 3 | 3 | 1 | 1 | 0 | 2 | 0 | 1 | 0 | 2 | 0 | 3 | 1 | 1 | 0 | 0 | 1 | 0 |
| Mermoz | 200 | 90 | 90 | 22.5 | 9 | 4 | 2 | 2 | 2 | 2 | 1 | 0 | 1 | 0 | 1 | 0 | 0 | 1 | 0 | 2 | 1 | 1 | 1 | 1 | 3 | 0 |
| FenetreMermoz | 240 | 96 | 96 | 19.2 | 10 | 5 | 2 | 3 | 2 | 2 | 1 | 0 | 1 | 2 | 0 | 1 | 0 | 1 | 0 | 2 | 1 | 1 | 1 | 1 | 2 | 0 |
| Foire | 280 | 350 | 270 | 33.8 | 15 | 8 | 3 | 6 | 2 | 3 | 1 | 1 | 1 | 2 | 1 | 1 | 0 | 2 | 1 | 3 | 0 | 0 | 0 | 1 | 2 | 0 |
| BelAir | 290 | 310 | 195 | 32.5 | 12 | 6 | 2 | 4 | 2 | 3 | 1 | 1 | 1 | 2 | 0 | 1 | 0 | 2 | 0 | 2 | 0 | 1 | 1 | 0 | 2 | 0 |
| SacreCoeur | 290 | 255 | 204 | 34.0 | 11 | 6 | 2 | 4 | 2 | 2 | 1 | 1 | 0 | 2 | 0 | 1 | 0 | 2 | 0 | 2 | 1 | 0 | 1 | 0 | 2 | 0 |
| Fass | 310 | 387 | 174 | 29.0 | 11 | 6 | 2 | 4 | 2 | 2 | 1 | 1 | 0 | 2 | 1 | 1 | 0 | 1 | 0 | 1 | 1 | 0 | 1 | 0 | 1 | 0 |
| Hann | 350 | 293 | 220 | 31.4 | 13 | 7 | 2 | 5 | 2 | 3 | 1 | 1 | 1 | 2 | 1 | 1 | 0 | 2 | 0 | 3 | 0 | 1 | 0 | 1 | 3 | 0 |
| Plateau | 350 | 60 | 60 | 15.0 | 8 | 4 | 1 | 3 | 1 | 2 | 1 | 0 | 1 | 2 | 0 | 1 | 0 | 0 | 0 | 0 | 1 | 0 | 1 | 1 | 3 | 1 |
| Mermoz | 390 | 203 | 198 | 33.0 | 11 | 6 | 2 | 4 | 2 | 2 | 1 | 1 | 0 | 2 | 1 | 1 | 0 | 2 | 1 | 2 | 1 | 0 | 1 | 1 | 2 | 0 |
| Fann | 440 | 100 | 100 | 20.0 | 10 | 5 | 2 | 3 | 2 | 2 | 1 | 0 | 1 | 3 | 1 | 1 | 0 | 1 | 0 | 2 | 1 | 1 | 1 | 1 | 3 | 0 |

| Plateau | 590 | 95 | 95 | 19.0 | 10 | 5 | 2 | 3 | 2 | 2 | 1 | 0 | 0 | 2 | 0 | 1 | 0 | 0 | 0 | 0 | 1 | 0 | 1 | 1 | 3 | 1 |
|---|---|---|---|---|---|---|---|---|---|---|---|---|---|---|---|---|---|---|---|---|---|---|---|---|---|---|
| PointE | 600 | 589 | 251 | 35.9 | 14 | 7 | 3 | 6 | 1 | 3 | 1 | 1 | 1 | 1 | 1 | 1 | 1 | 2 | 0 | 2 | 1 | 0 | 1 | 1 | 3 | 0 |
| PointE | 690 | 250 | 304 | 30.4 | 17 | 10 | 3 | 7 | 3 | 3 | 1 | 1 | 1 | 2 | 0 | 1 | 0 | 3 | 1 | 2 | 1 | 0 | 1 | 1 | 3 | 0 |
| Ngor | 800 | 307 | 285 | 31.7 | 15 | 9 | 2 | 6 | 3 | 3 | 1 | 1 | 1 | 3 | 1 | 1 | 1 | 2 | 0 | 3 | 1 | 1 | 1 | 1 | 3 | 0 |
| FannHock | 840 | 252 | 270 | 24.5 | 18 | 11 | 3 | 8 | 3 | 3 | 1 | 1 | 1 | 2 | 0 | 1 | 0 | 3 | 0 | 2 | 1 | 1 | 1 | 1 | 3 | 0 |
| Mamelles | 850 | 598 | 294 | 29.4 | 17 | 10 | 3 | 6 | 4 | 2 | 2 | 1 | 1 | 3 | 1 | 1 | 0 | 2 | 1 | 3 | 0 | 1 | 1 | 1 | 3 | 0 |
| Mamelles | 970 | 396 | 396 | 39.6 | 18 | 10 | 3 | 7 | 3 | 4 | 1 | 1 | 1 | 2 | 0 | 1 | 1 | 3 | 1 | 3 | 0 | 1 | 1 | 1 | 3 | 0 |
| Almadies | 1000 | 500 | 346 | 28.8 | 20 | 12 | 3 | 8 | 4 | 4 | 1 | 1 | 1 | 3 | 1 | 1 | 1 | 3 | 2 | 3 | 0 | 1 | 1 | 1 | 3 | 0 |
| Plateau | 1370 | 154 | 154 | 25.7 | 12 | 6 | 2 | 4 | 2 | 3 | 1 | 1 | 1 | 2 | 0 | 1 | 0 | 2 | 0 | 0 | 1 | 0 | 1 | 1 | 3 | 1 |
| FannResidence | 2100 | 988 | 390 | 35.5 | 20 | 11 | 4 | 8 | 3 | 4 | 1 | 1 | 1 | 3 | 1 | 1 | 0 | 3 | 2 | 2 | 1 | 1 | 1 | 1 | 3 | 0 |